\documentclass[10pt,journal,compsoc]{IEEEtran}
\IEEEoverridecommandlockouts

\usepackage[dvipsnames, table, svgnames]{xcolor}
\usepackage{tikz}
\definecolor{hidden-draw}{RGB}{128, 128, 128}

\usepackage{amsmath,amssymb,amsfonts,bm}
\usepackage{graphicx}%
\usepackage{multirow}%
\usepackage{mathrsfs}%
\usepackage{textcomp}%
\usepackage{manyfoot}%
\usepackage{booktabs}%
\usepackage{algorithm}%
\usepackage{algorithmicx}%
\usepackage{algpseudocode}%
\usepackage{listings}%
\usepackage{longtable}
\usepackage{supertabular}
\usepackage{enumitem}
\usepackage[graphicx]{realboxes}
\usepackage{balance}
\usepackage{textcomp}
\usepackage[table]{xcolor}
\usepackage{subfigure}
\usepackage{url}
\usepackage{booktabs}
\usepackage{array}
\usepackage{caption}
\usepackage[T1]{fontenc}
\usepackage[colorlinks, linkcolor=blue, citecolor=blue, urlcolor=blue]{hyperref}
\usepackage{color}
\usepackage[edges]{forest}

\def\BibTeX{{\rm B\kern-.05em{\sc i\kern-.025em b}\kern-.08em
    T\kern-.1667em\lower.7ex\hbox{E}\kern-.125emX}}

\begin{document}
\definecolor{lightred}{rgb}{1, 0.92, 0.92}
\definecolor{lightorange}{rgb}{1, 0.95, 0.878}
\definecolor{lightgray}{gray}{0.9}
\definecolor{lightyellow}{rgb}{1, 1, 0.839}
\definecolor{lightgreen}{rgb}{0.901, 1, 0.882}
\definecolor{lightblue}{rgb}{0.901, 1, 1}
\definecolor{lightcyan}{rgb}{0.901, 0.941, 1}
\definecolor{lightpurple}{rgb}{0.960, 0.858, 0.996}
\newcommand{\hx}[1]{{{\textcolor{cyan}{[Haixin: #1]}}}}
\newcommand{\yadi}[1]{{{\textcolor{cyan}{[Yadi: #1]}}}}
\newcommand{\yuxuan}[1]{{{\textcolor{magenta}{[Yuxuan: #1]}}}}
\newcommand{\XL}[1]{{\bf\color{purple}[{\sc Xiao:} #1]}}
\newcommand{\ZH}[1]{{\bf\color{orange}[{\sc Zijie:} #1]}}
\newcommand{\YS}[1]{{\bf\color{red}[{\sc YS:} #1]}}

\title{Recent Advances on Machine Learning for
Computational Fluid Dynamics: A Survey}

\author{
~Haixin~Wang,~Yadi~Cao,~Zijie~Huang,~Yuxuan~Liu,~Peiyan~Hu,~Xiao~Luo,~Zezheng~Song,~Wanjia \\ Zhao,~Jilin~Liu,~Jinan~Sun\IEEEauthorrefmark{2},~Shikun~Zhang,~Long~Wei,~Yue~Wang,~Tailin~Wu,~Zhi-Ming~Ma,~Yizhou~Sun

\IEEEcompsocitemizethanks{

\IEEEcompsocthanksitem H. Wang, J. Liu, J. Sun and S. Zhang are with Peking University.
E-mail: wang.hx@stu.pku.edu.cn, sjn@pku.edu.cn.

\IEEEcompsocthanksitem 
Y. Cao, Z. Huang, X. Luo, Y. Liu and Y. Sun are with University of California, Los Angeles.

\IEEEcompsocthanksitem 
P. Hu and Z. Ma are from Academy of Mathematics and Systems Science, Chinese Academy of Sciences.

\IEEEcompsocthanksitem 
Z. Song is from University of Maryland, College Park.

\IEEEcompsocthanksitem 
W. Zhao is from Stanford University.

\IEEEcompsocthanksitem 
L. Wei and T. Wu work at Westlake University. 

\IEEEcompsocthanksitem 
Y. Wang works at Microsoft AI4Science.

\IEEEcompsocthanksitem
\IEEEauthorrefmark{2}Corresponding author.
}
}

\IEEEtitleabstractindextext{%
\begin{abstract}
This paper explores the recent advancements in enhancing Computational Fluid Dynamics (CFD) tasks through Machine Learning (ML) techniques. 
We begin by introducing fundamental concepts, traditional methods, and benchmark datasets, then examine the various roles ML plays in improving CFD. The literature systematically reviews papers in recent five years and introduces a novel classification for forward modeling: Data-driven Surrogates, Physics-Informed Surrogates, and ML-assisted Numerical Solutions. 
Furthermore, we also review the latest ML methods in inverse design and control, offering a novel classification and providing an in-depth discussion.
Then we highlight real-world applications of ML for CFD in critical scientific and engineering disciplines, including aerodynamics, combustion, atmosphere \& ocean science, biology fluid, plasma, symbolic regression, and reduced order modeling. 
Besides, we identify key challenges and advocate for future research directions to address these challenges, such as multi-scale representation, physical knowledge encoding, scientific foundation
model and automatic scientific discovery.
This review serves as a guide for the rapidly expanding ML for CFD community, aiming to inspire insights for future advancements. 
We draw the conclusion that ML is poised to significantly transform CFD research by enhancing simulation accuracy, reducing computational time, and enabling more complex analyses of fluid dynamics.
The paper resources can be viewed at \url{https://github.com/WillDreamer/Awesome-AI4CFD}. 
\end{abstract}

}

\maketitle

\IEEEdisplaynontitleabstractindextext
\IEEEpeerreviewmaketitle

\IEEEraisesectionheading{\section{Introduction}}

\IEEEPARstart{F}LUID dynamics is a fundamental discipline that studies the motion and behavior of fluid flow. It serves as a foundation across a wide range of scientific and engineering fields, including aerodynamics~\cite{o2022neural,deng2023prediction,mufti2024shock}, chemical engineering~\cite{cao2018liquid,cao2019liquid,zhang2022multiscale}, biology~\cite{yin2022simulating,voorter2023improving,shen2023multiple}, and environmental science~\cite{wen2022u,pathak2022fourcastnet,bi2023accurate,jiang2023fourier,lam2023learning,rajagopal2023evaluation}. CFD 
employs mathematical models to simulate fluid dynamics through partial differential equations (PDEs)~\cite{versteeg2007introduction}. The primary goal of CFD is to obtain simulated results under various working conditions, thereby reducing the need for costly real-world experiments and accelerating engineering design and control processes.

\begin{figure}[t]
\centering
\includegraphics[width=0.45\textwidth]{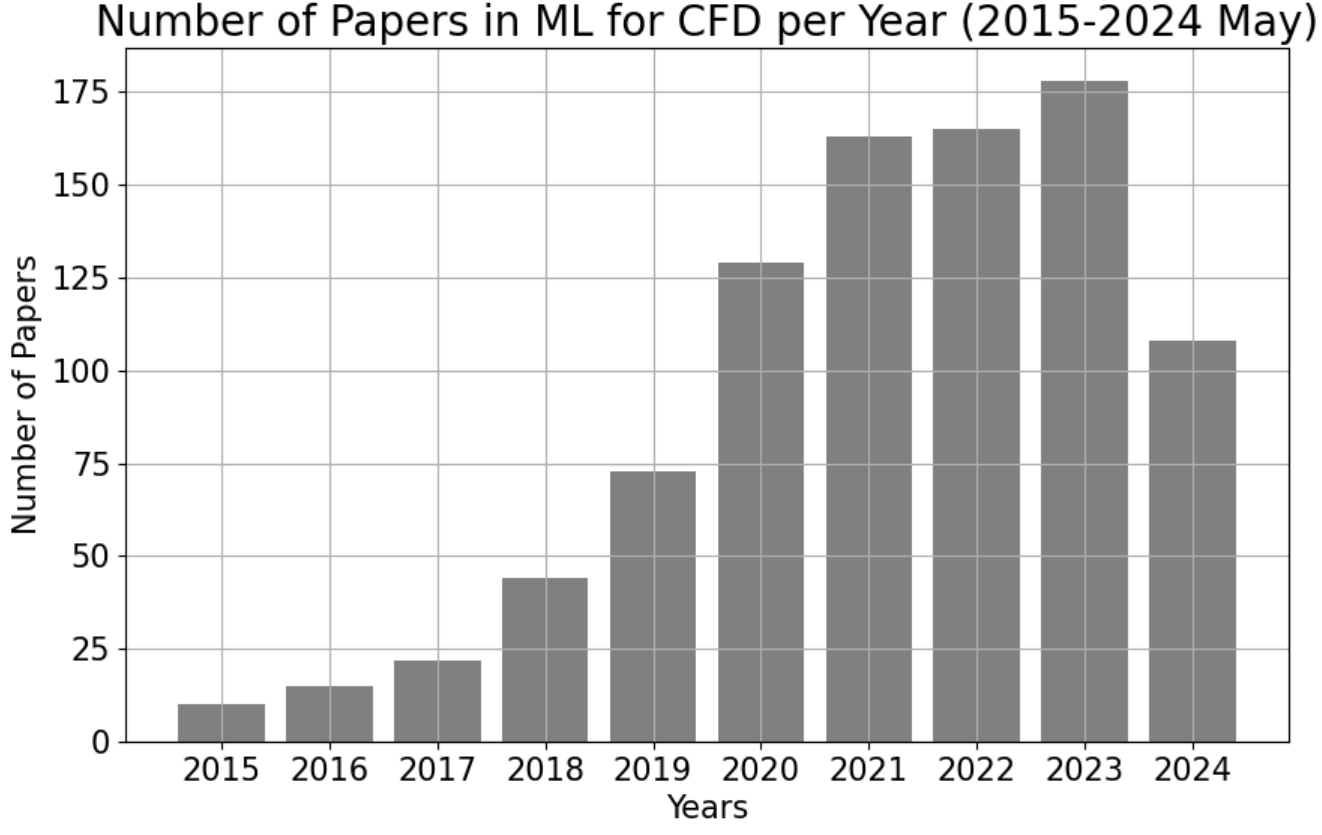}
\vspace{-0.3cm}
\caption{The approximate annual number of papers on ML for CFD presented at top-tier ML publication and leading journals in fluid dynamics appeared in Table \ref{tab:method} and \ref{tab:method_physics}}
\label{fig:paper}
\vspace{-0.5cm}
\end{figure}

Despite decades of advancement in research and engineering practice, CFD techniques continue to face significant challenges. These include high computational costs due to demanding restrictions on spatial or temporal resolutions, difficulties in capturing subscale dynamics such as in turbulence~\cite{zhang2023review}, and stability issues with numerical algorithms~\cite{versteeg2007introduction}, among others.
On the other hand, ML, famous for its ability to learn patterns and dynamics from observed data, has recently emerged as a trend that can reshape or enhance any general scientific subject~\cite{thiyagalingam2022scientific}. The integration of ML techniques with the extensive fluid dynamics data accumulated over recent decades offers a transformative approach to augment CFD practices (see Fig. \ref{fig:paper}). As the field of ML continues to expand rapidly, it becomes increasingly challenging for researchers to stay updated. In response, this review aims to shed light on the multifaceted roles ML plays in enhancing CFD.

Actually, there have already been some surveys on the application of ML methods in the CFD field. However, most of these surveys have the following two limitations: 1) \textbf{Only earlier attempts.} For instance, Wang \textit{et al.} \cite{wang2021physics} and Huang \textit{et al.} \cite{huang2022partial} both provide a detailed discussion on incorporating physics-based modeling into ML, emphasizing dynamic systems and hybrid approaches.
Similarly, Vinuesa \textit{et al.} \cite{vinuesa2022enhancing} explores promising ML directions from the perspective of CFD domain, such as direct numerical simulations, a large-eddy simulation (LES), a schematic of the turbulence spectrum, Reynolds-averaged Navier–Stokes (RANS) simulations, and a dimensionality-reduction method. However, they only review early ML applications to PDEs, with a focus on works prior to 2021.  
2) \textbf{Incomplete overview}. The current body of surveys on ML applications in CFD primarily focuses on integrating physical knowledge and common model architectures for PDEs. Zhang \textit{et al.} \cite{zhang2023artificial} examine ML for both forward and inverse modeling of PDEs, highlighting four key challenges but ignoring systematic classification and the potential applications in this area. Meanwhile, Lino \textit{et al.} \cite{lino2023current} roughly differentiate between physics-driven and data-driven approaches and address several methodological limitations, but similarly overlook systematic classification on the motivation of each method.

\begin{figure*}[t]
\centering
\includegraphics[width=0.96\textwidth]{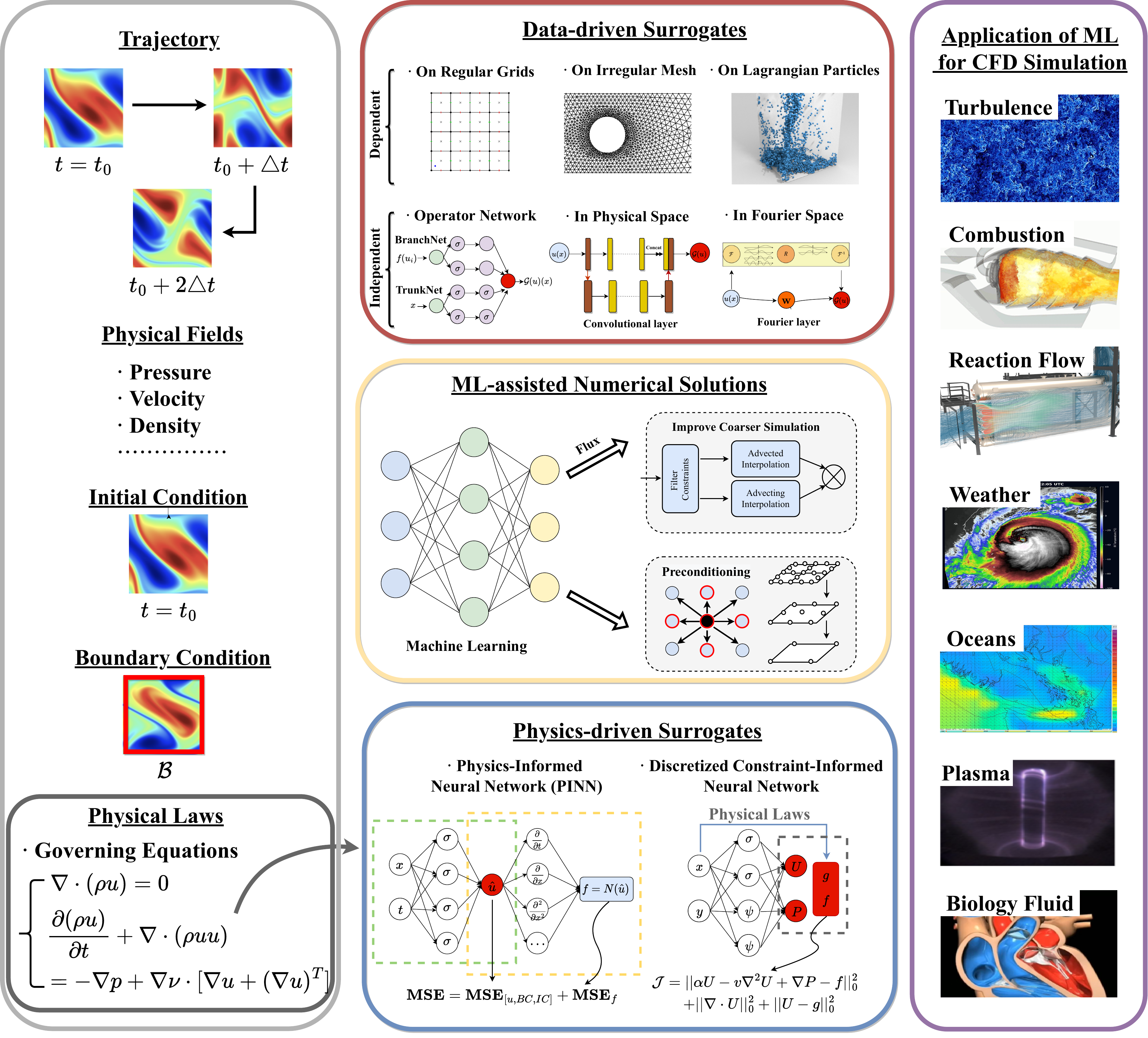}
\vspace{-0.3cm}
\caption{Overview of ML for computational fluid dynamics simulation. 
\textbf{The left column} encompasses various types of input data used in the models, including physical laws. \textbf{The middle columns} consist of three common frameworks used in constructing models with ML. \textbf{The right column} pertains to applications in various scenarios.
}
\label{fig:framework}
\vspace{-0.5cm}
\end{figure*}

Despite these contributions, there remains a gap in comprehensive, cutting-edge, and profound systematization of ML methods for CFD. Our work represents the first survey that consolidates these fragmented insights into a cohesive framework. We systematically review the fundamental knowledge, data, methodologies, applications, challenges and future directions in the field.
The structure of this paper is shown in Fig. \ref{fig_tax} and organized as follows: 

In Section~\ref{sec:pre}, we introduce the fundamental concepts and knowledge of CFD, accompanied by an annotated list of all the types of PDEs addressed by the literature reviewed.
We then systematically review the literature from the recent five years, categorizing the selected studies into three primary disciplines with demonstration in Fig. \ref{fig:framework}: \textbf{Data-driven Surrogates} (Section~\ref{sec:data_driven_surrogates}), which rely exclusively on observed data for training; \textbf{Physics-Informed Surrogates} (Section~\ref{sec:physics_driven_surrogates}), which integrate selected physics-informed priors into ML modeling; and \textbf{ML-assisted Numerical Solutions} (Section~\ref{sec:ml_assisted_solutions}), which partially replace traditional numerical solvers to achieve a balance between efficiency, accuracy, and generalization. 
In addition, we introduce the setup of inverse design and control problems (Section \ref{sec:lab}), which are two fundamental problems when applying CFD to real-world applications. The former optimizes the design parameters, e.g., initial and boundary conditions, for certain design objectives. And the latter is to control a physical system to achieve a specific objective by applying time-varying external forces.

Following this, Section~\ref{sec:app} discusses the application of these methods across key scientific and engineering disciplines, showcasing their impact and potential.
Finally, Section~\ref{sec:chall} addresses the key challenges and limitations within the current state-of-the-art and outlines prospective research directions. 
We aim to draw the attention of the broader ML community to this review, enriching their research with fundamental CFD knowledge and advanced developments, thus inspiring future work in the field. 

\noindent\textbf{Differences from Existing Surveys. } Compared with existing surveys, our survey has four inclusive features: 
(1) \underline{\textit{Update-to-date Summarization}}. This survey focuses on the newest papers from 2020 to 2024 based on the current state of development. In contrast, existing related surveys were published before 2022. 
(2) \underline{\textit{Innovative Taxonomy.}} This survey systematically reviews the ML methods in the CFD field and introduces a novel classification based on the motivations behind methods designed for both forward modeling and inverse problems for the first time.
(3) \underline{\textit{Comprehensive Discussion}}. This survey provides a comprehensive discussion, covering background, data, forward modeling/inverse design methods, and applications, which helps researchers quickly and thoroughly understand this field.
(4) \underline{\textit{Future Guidance.}} Our work summarizes the most recent advancements in CFD and highlights the challenges in current CFD research, which can provide guidance and direction for future work in the field, \textit{i.e.,} scientific foundation model.

\noindent\textbf{Broader Impact.} The impact of our survey lies in two points. (1) \underline{\textit{To Science-related Community.}} Our survey summarizes effective ML approaches for CFD, which can help researchers in physics and mechanics find solutions and benefits from ML. (2) \underline{\textit{To ML Community.}} Our survey can also provide guidance for ML researchers and help them apply their knowledge to real-world scientific applications in CFD.

\tikzstyle{my-box}=[
    rectangle,
    draw=hidden-draw,
    rounded corners,
    align=left,
    text opacity=1,
    minimum height=1.5em,
    minimum width=5em,
    inner sep=2pt,
    fill opacity=.8,
    line width=0.8pt,
]

\tikzstyle{leaf-head}=[my-box, minimum height=1.5em,
    draw=gray!80, 
    fill=gray!35,  
    text=black, font=\normalsize,
    inner xsep=2pt,
    inner ysep=4pt,
    line width=0.8pt,
]

\tikzstyle{leaf-task}=[my-box, minimum height=2.5em,
    draw=red!80, 
    fill=red!20,  
    text=black, font=\normalsize,
    inner xsep=2pt,
    inner ysep=4pt,
    line width=0.8pt,
]

\tikzstyle{leaf-paradigms}=[my-box, minimum height=2.5em,
    draw=orange!70, 
    fill=orange!15,  
    text=black, font=\normalsize,
    inner xsep=2pt,
    inner ysep=4pt,
    line width=0.8pt,
]
\tikzstyle{leaf-others}=[my-box, minimum height=2.5em,
    draw=yellow!80, 
    fill=yellow!15,  
    text=black, font=\normalsize,
    inner xsep=2pt,
    inner ysep=4pt,
    line width=0.8pt,
]
\tikzstyle{leaf-other}=[my-box, minimum height=2.5em,
    draw=green!80, 
    fill=green!15,  
    text=black, font=\normalsize,
    inner xsep=2pt,
    inner ysep=4pt,
    line width=0.8pt,
]
\tikzstyle{leaf-application}=[my-box, minimum height=2.5em,
    draw=blue!80, 
    fill=blue!15,  
    text=black, font=\normalsize,
    inner xsep=2pt,
    inner ysep=4pt,
    line width=0.8pt,
]

\tikzstyle{modelnode-task}=[my-box, minimum height=1.5em,
    draw=red!80, 
    fill=red!20,  
    text=black, font=\normalsize,
    inner xsep=2pt,
    inner ysep=4pt,
    line width=0.8pt,
]

\tikzstyle{modelnode-paradigms}=[my-box, minimum height=1.5em,
    draw=orange!70, 
    fill=orange!15,  
    text=black, font=\normalsize,
    inner xsep=2pt,
    inner ysep=4pt,
    line width=0.8pt,
]
\tikzstyle{modelnode-others}=[my-box, minimum height=1.5em,
    draw=yellow!80, 
    fill=yellow!15,  
    text=black, font=\normalsize,
    inner xsep=2pt,
    inner ysep=4pt,
    line width=0.8pt,
]
\tikzstyle{modelnode-other}=[my-box, minimum height=1.5em,
    draw=green!80, 
    fill=green!15,  
    text=black, font=\normalsize,
    inner xsep=2pt,
    inner ysep=4pt,
    line width=0.8pt,
]
\tikzstyle{modelnode-application}=[my-box, minimum height=1.5em,
    draw=blue!80, 
    fill=blue!15,  
    text=black, font=\normalsize,
    inner xsep=2pt,
    inner ysep=4pt,
    line width=0.8pt,
]
\begin{figure*}[!th]
    \centering
    \resizebox{1\textwidth}{!}
    {
        \begin{forest}
            for tree={
                grow=east,
                reversed=true,
                anchor=base west,
                parent anchor=east,
                child anchor=west,
                base=left,
                font=\normalsize,
                rectangle,
                draw=hidden-draw,
                rounded corners,
                align=left,
                minimum width=1em,
                edge+={darkgray, line width=1pt},
                s sep=3pt,
                inner xsep=0pt,
                inner ysep=3pt,
                line width=0.8pt,
                ver/.style={rotate=90, child anchor=north, parent anchor=south, anchor=center},
            }, 
            [
                Machine Learning for Computational Fluid Dynamics,leaf-head, ver 
                [    
                    Forward Modeling,leaf-head, ver 
                    [
                     \S \ref{sec:data_driven_surrogates} Data-driven \\ Surrogates, leaf-task,text width=8em
                    [
                        \S \ref{subsec:dependent_discretization} Dependent \\ on Discretization, leaf-task, text width=8em
                        [
                            \S \ref{subsubsec:structured_grids} Structu-\\red Grids, leaf-task, text width=7em
                            [\textbf{DPUF}~\cite{lee2019data}{, } \textbf{TF-Net}~\cite{wang2020towards}{, } \textbf{EquNet}~\cite{wang2020incorporating}{, } \textbf{RSteer}~\cite{wang2022approximately}, modelnode-task, text width=39.5em]
                        ]
                        [
                            \S \ref{subsubsec:unstructured_mesh} Unstru-\\ctured Mesh, leaf-task, text width=7em
                            [\textbf{GNS}~\cite{sanchez2020learning}{, } \textbf{MGN}~\cite{pfaff2020learning}{, } \textbf{MP-PDE}~\cite{brandstetter2022message}{, } \textbf{Han \textit{et al.}}~\cite{han2022predicting}{, } \textbf{TIE}~\cite{shao2022transformer}{, } \textbf{MAgNet}~\cite{boussif2022magnet}{, } \\ \textbf{GNODE}~\cite{bishnoi2022enhancing}{, } \textbf{FCN}~\cite{he2022flow}{, } \textbf{Zhao \textit{et al.}}~\cite{zhao2023computationally}{, } \textbf{DINo~}\cite{yin2022continuous}{, } \textbf{LAMP}~\cite{wu2023learning}{, }
                            \textbf{CARE}~\cite{luo2023care}{, }
                            \\
                            \textbf{BENO}~\cite{wang2024beno}{, }
                            \textbf{HAMLET}~\cite{bryutkin2024hamlet}
                            , modelnode-task, text width=39.5em]
                        ]
                        [
                            \S \ref{subsubsec:lagrangian_particles} Lagran-\\gian Particles, leaf-task, text width=7em
                            [\textbf{CC}~\cite{ummenhofer2019lagrangian}{, } \textbf{Wessels \textit{et al.}}~\cite{wessels2020neural}{, } \textbf{FGN}~\cite{li2022graph}{, } \textbf{MCC}~\cite{liu2023fast}{, } \textbf{LFlows}~\cite{torres2023lagrangian}{, } \textbf{Li \textit{et al.}}~\cite{li2024synthetic}, modelnode-task, text width=39.5em]
                        ]
                    ]
                    [
                        \S \ref{subsec:independent_discretization} Independent \\ on Discretization, leaf-task, text width=8em
                         [
                            \S \ref{sec:don} Deep Op-\\erator Network, leaf-task, text width=7em
                            [\textbf{DeepONet}~\cite{lu2021learning}{,} \textbf{PI-DeepONet}~\cite{wang2021learning}{,} \textbf{MIONet}~\cite{jin2022mionet}{,} \textbf{B-DeepONet}~\cite{lin2023b}{,} \textbf{NOMAD}~\cite{seidman2022nomad}{,} \\
                            \textbf{Fourier-MIONet}~\cite{jiang2023fourier}{,} \textbf{Shift-DeepONet}~\cite{lanthaler2022nonlinear}{,} \textbf{HyperDeepONet}~\cite{lee2023hyperdeeponet}{,} \textbf{L-DeepONet}~\cite{kontolati2023learning}{,} \\ \textbf{SVD-DeepONet}~\cite{venturi2023svd}, modelnode-task, text width=39.5em]
                        ]
                        [
                            \S \ref{subsubsec:other_operator_learning} In Phys-\\ical Space, leaf-task, text width=7em [\textbf{MGNO}~\cite{li2020multipole}{,} \textbf{G. \textit{et al.}}~\cite{geneva2022transformers}{,} \textbf{GNOT}~\cite{hao2023gnot}{,} \textbf{CNO}~\cite{raonic2023convolutional}{,} \textbf{LNO}~\cite{cao2023lno}{,} \textbf{KNO}~\cite{xiong2023koopman}{,} \textbf{ICON}~\cite{yang2023context}, modelnode-task, text width=39.5em]
                        ]
                        [
                            \S \ref{subsubsec:f_operator_learning} In Four-\\ier Space, leaf-task, text width=7em
                            [\textbf{FNO}~\cite{li2020fourier}{,} \textbf{PINO}~\cite{li2021physics}{,} \textbf{GEO-FNO}~\cite{li2022fourier}{,} \textbf{U-NO}~\cite{rahman2022u}{,} \textbf{SSNO}~\cite{rafiq2022ssno}{,} \textbf{F-FNO}~\cite{tran2022factorized}{,} \textbf{CFNO}~\cite{brandstetter2022clifford}{,}\\ \textbf{CMWNO}~\cite{xiao2022coupled}{,} \textbf{MCNP}~\cite{zhang2023monte}{,} \textbf{$G$-FNO}~\cite{helwig2023group}{,} \textbf{GINO}~\cite{li2023geometry}{,} \textbf{DAFNO}~\cite{liu2023domain}{,} \textbf{FCG-NO}~\cite{rudikov2024neural}, modelnode-task, text width=39.5em]
                        ]
                    ]
                ]
                [
                    \S \ref{sec:physics_driven_surrogates} Physics-driven \\ Surrogates, leaf-paradigms,text width=8em
                    [
                        \S \ref{subsec:pinn} PINNs, leaf-paradigms, text width=9em
                        [\textbf{PINN}~\cite{raissi2019physics}{, }\textbf{Bai \textit{et al.}}~\cite{bai2020ApplyingPhysicsInformed}{, } \textbf{PINN-SR}~\cite{chen2021physics}{, } \textbf{NSFnet}~\cite{jin_nsfnets_2021}{, } \textbf{Phygeonet}~\cite{gao2021phygeonet}{, } \textbf{PINN-LS}~\cite{mowlavi2021optimal}{, }  \textbf{ResPINN}~\cite{cheng2021deep}{,}\\\textbf{CPINN}~\cite{zeng2022competitive}{, } \textbf{stan}~\cite{gnanasambandam2022self}{, } \textbf{Meta-Auto-Decoder}~\cite{huang2022meta}{, } \textbf{BINN}~\cite{sun2023binn}{, } \textbf{Hodd-PINN}~\cite{you2023high}{, } \textbf{DATS}~\cite{toloubidokhti2023dats}{, } \\\textbf{PINNsFormer}~\cite{zhao2023pinnsformer}{, } \textbf{PeRCNN}~\cite{rao2023encoding}{, } \textbf{NASPINN}~\cite{wang2024pinn},  modelnode-paradigms, text width=47.5em]
                    ]
                    [
                       \S \ref{subsec:discritized_pde} Constraint-Informed, leaf-paradigms, text width=9em
                         [\textbf{DGM}\cite{sirignano2018dgm}{, }\textbf{AmorFEA}\cite{xue2020amortized}{, }\textbf{EDNN}\cite{du2021evolutional}{, }\textbf{Liu \textit{et al.}}\cite{liu2022predicting}{, } \textbf{Gao \textit{et al.}}~\cite{gao2022physics}{, }\textbf{ADLGM \textit{et al.}}~\cite{aristotelous2023adlgm}{, } \textbf{INSR}~\cite{chen2023implicit}{,}\\\textbf{Neu-Gal}~\cite{bruna2024neural}{, } \textbf{PPNN}\cite{liu2024multi}{, } \textbf{FEX}~\cite{song2024finite},                          modelnode-paradigms, text width=47.5em]
                    ]
                ]
                [
                    \S \ref{sec:ml_assisted_solutions} ML-Assisted \\Numerical \\ Solutions, leaf-others,text width=7.5em
                    [
                        \S \ref{subsec:discretization_scheme} Coarser Scales, leaf-others, text width=9em
                        [\textbf{Zhang \textit{et al.}}~\cite{zhang2022multiscale}{, }\textbf{Kochkov \textit{et al.}}~\cite{kochkov2021machine}{, }\textbf{Despres \textit{et al.} }~\cite{despres2020machine}{, } \textbf{Bar-Sinai \textit{et al.}}~\cite{bar2019learning}{, }\\ \textbf{List \textit{et al.}}~\cite{list2022learned}{, }\textbf{Sun \textit{et al.}}~\cite{sun2023neural}, modelnode-others, text width=48em]
                    ]
                    [
                        \S \ref{subsec:multi_grid} Preconditioning, leaf-others, text width=9em
                         [\textbf{Greenfeld \textit{et al.}}~\cite{greenfeld2019learning}{, }\textbf{Luz \textit{et al.}}~\cite{luz2020learning}{, }\textbf{Sappl\textit{et al.}}~\cite{sappl2019deep}, modelnode-others, text width=48em]
                    ]
                    [
                        \S \ref{subsec:adhoc_combine} Miscellaneous, leaf-others, text width=9em
                        [\textbf{Pathak \textit{et al.}}~\cite{pathak2020using}{, }\textbf{Obiols \textit{et al.}}~\cite{obiols2020cfdnet}{, }\textbf{Um \textit{et al.}}~\cite{um2020solver}{, }\textbf{CFD-GCN}~\cite{belbute2020combining}, modelnode-others,text width=48em]
                    ]
                ]
                ]
                [
                    \S \ref{sec:lab} Inverse \\ Design \& Control, leaf-other,text width=8em
                    [
                        \S \ref{sec:inverse_design} Inverse Design, leaf-other, text width=9em
                        [
                            \S \ref{sec:PDE-constrained} PDE-constrained, leaf-other, text width=11em
                            [\textbf{hPINN}~\cite{lu2021physics}{, }\textbf{gPINN}~\cite{yuGradientenhancedPhysicsinformedNeural2022}{, }\textbf{Bi-PINN}~\cite{hao2022bi}{, }\textbf{Pokkunuru \textit{et al.}}~\cite{pokkunuru2022improved}, modelnode-other, text width=37.8em]
                        ]
                        [
                            \S \ref{sec:Data_driven_inverse} Data-driven, leaf-other, text width=11em
                            [\textbf{Allen \textit{et al.}}~\cite{allen2022inverse}{, }\textbf{Wu \textit{et al.}}~\cite{wu2022learning}{, } \textbf{Ardizzone \textit{et al.}}~\cite{ardizzone2018analyzing}{, } \textbf{INN}~\cite{behrmann2019invertible}{, } \\ 
                            \textbf{Invertible AE}~\cite{teng2019invertible}{, } \textbf{cINN}~\cite{kruse2021benchmarking}{, } \textbf{Ren \textit{et al.}}~\cite{ren2020benchmarking}{, } \textbf{Kumar \textit{et al.}}~\cite{kumar2020model}, modelnode-other, text width=37.8em]
                        ]
                    ]
                    [
                        \S \ref{sec:control} Control, leaf-other, text width=9em
                        [
                            \S \ref{sec:contril_sl} Supervised Learning, leaf-other, text width=13.5em
                            [\textbf{Holl \textit{et al.}}~\cite{holl2020learning}{, }\textbf{Hwang \textit{et al.}}~\cite{hwang2022solving}, modelnode-other, text width=35.4em]
                        ]
                        [
                            \S \ref{sec:contril_rl} Reinforcement Learning, leaf-other, text width=13.5em
                            [\textbf{Viquerat \textit{et al.}}~\cite{viquerat2022review}{, }\textbf{Garnier \textit{et al.}}~\cite{garnier2021review}{, }\textbf{Garnier \textit{et al.}}~\cite{GARNIER2021104973}, modelnode-other, text width=35.4em]
                        ]
                        [
                            \S \ref{sec:contril_PDE-Constrained} PDE-constrained, leaf-other, text width=13.5em
                            [\textbf{Mowlavi \textit{et al.}}\cite{MOWLAVI2023111731}{, }\textbf{Barrystraume \textit{et al.}}~\cite{barrystraume2022physicsinformed}, modelnode-other, text width=35.4em]
                        ]
                    ]
                ]
                [
                    \S \ref{sec:app} Applications, leaf-application,text width=8em
                    [
                        \S \ref{sub:Aerodynamics} Aerodynamics, leaf-application, text width=16em
                        [\textbf{Mao \textit{et al.}}~\cite{mao2020physics}{, }\textbf{Huang \textit{et al.}}~\cite{huang2022direct}{, }\textbf{Sharma \textit{et al.}}~\cite{sharma2023physics}{, }\textbf{Auddy \textit{et al.}}~\cite{auddy2023grinn}{, } \textbf{Shan \textit{et al.}}~\cite{shan2023turbulence}{, } \\ \textbf{Deng \textit{et al.}}~\cite{deng2023prediction}{, } \textbf{Mufti \textit{et al.}}~\cite{mufti2024shock}, modelnode-application, text width=43.5em]
                    ]
                    [
                        \S \ref{sub:Combustion} Combustion \& Reacting Flow, leaf-application, text width=16em
                         [\textbf{Ji \textit{et al.}}~\cite{Ji_2021}{, }\textbf{Zhang \textit{et al.}}~\cite{zhang2022multiscale}, modelnode-application, text width=43.5em]
                    ]
                    [
                        \S \ref{sub:Atmosphere} Atmosphere \& Ocean Science, leaf-application, text width=16em
                        [\textbf{Pathak \textit{et al.}}~\cite{pathak2022fourcastnet}{, }\textbf{Bi \textit{et al.}}~\cite{bi2023accurate}{, }\textbf{Lam \textit{et al.}}~\cite{lam2023learning}{, }\textbf{Rajagopal \textit{et al.}}~\cite{rajagopal2023evaluation}{, } \textbf{Jiang \textit{et al.}}~\cite{jiang2023fourier}, modelnode-application,text width=43.5em]
                    ]
                    [
                        \S \ref{sub:Biology} Biology Fluid, leaf-application, text width=16em
                        [\textbf{Yin \textit{et al.}}~\cite{yin2022simulating}{, }\textbf{Voorter \textit{et al.}}~\cite{voorter2023improving}{, }\textbf{Shen \textit{et al.}}~\cite{shen2023multiple}, modelnode-application,text width=43.5em]
                    ]
                    [
                        \S \ref{sub:Plasma} Plasma, leaf-application, text width=16em
                        [\textbf{Zhong \textit{et al.}}~\cite{Zhong_2022}{, }\textbf{Gopakumar \textit{et al.}}~\cite{gopakumar2023fourier}{, } \textbf{Kim \textit{et al.}}~\cite{kim2024highest}, modelnode-application,text width=43.5em]
                    ]
                    [
                        \S \ref{sub:symbolic} Symbolic Regression, leaf-application, text width=16em
                        [\textbf{FEX}~\cite{jiang2023finite}{, }\textbf{Becker \textit{et al.}}~\cite{becker2023predicting}, modelnode-application,text width=43.5em]
                    ]
                    [
                        \S \ref{sub:ROMs} Reduced Order Modeling, leaf-application, text width=16em
                        [\textbf{Leask \textit{et al.}}~\cite{leask2021modal}{, }\textbf{Geneva \textit{et al.}}~\cite{geneva2022transformers}{, }\textbf{Kneer}~\cite{kneer2021symmetry}{, }\textbf{Arnold \textit{et al.}}~\cite{arnold2022large}{, }\textbf{Wentland \textit{et al.}}~\cite{wentland2023scalable}, modelnode-application,text width=43.5em]
                    ]
                ]
            ]
        \end{forest}
    }
    \caption{Taxonomy of CFD methods based on ML techniques. We first investigate into forward modeling approaches, including data-driven surrogates, physics-driven surrogates, and ML-assisted methods. Besides, we conduct an in-depth analysis of inverse problems. Moreover, we review the practical applications of these methods across various domains.
    }
    \label{fig_tax}
\vspace{-0.3cm}
\end{figure*}
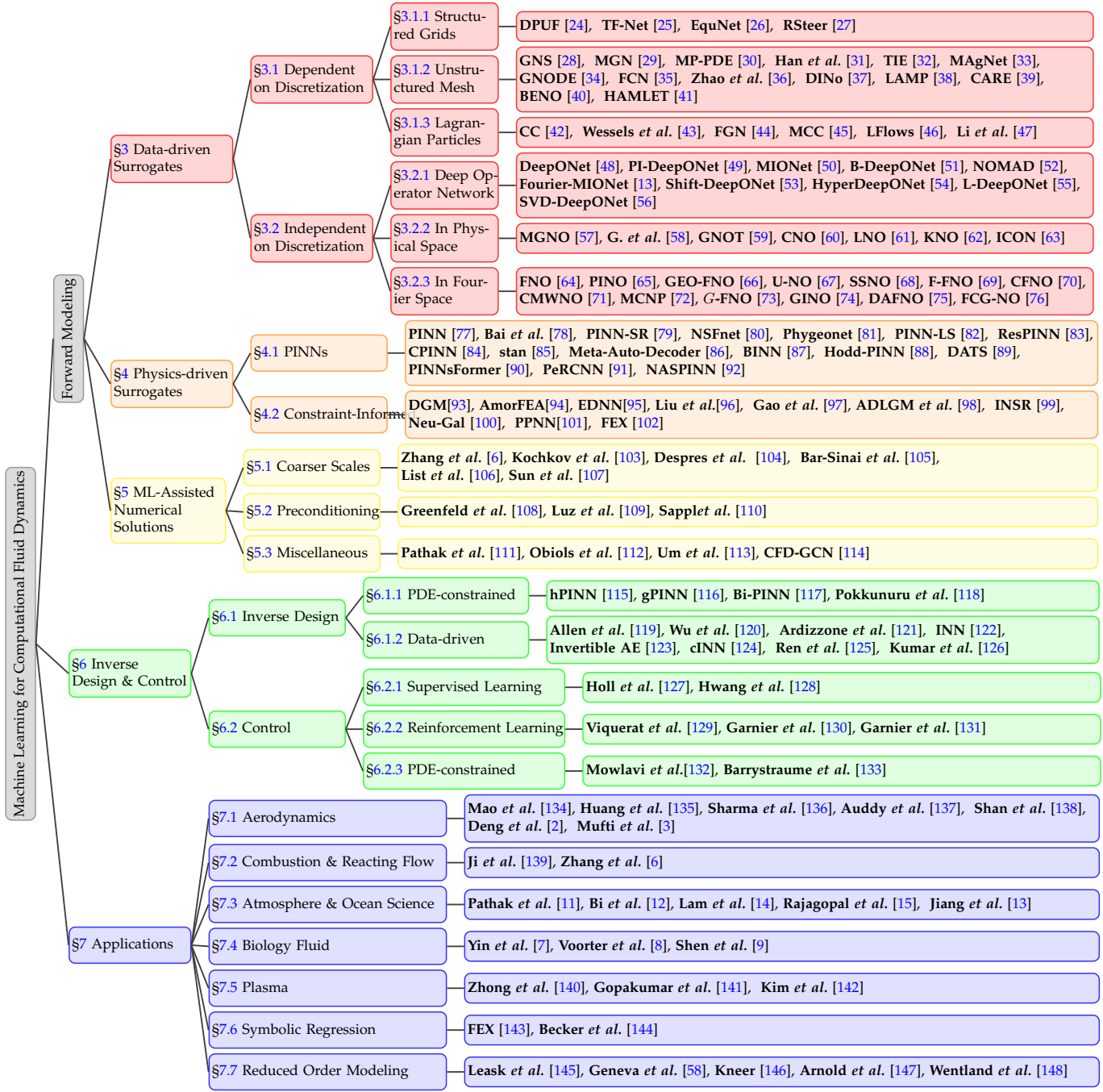

\begin{figure*}[t]
\centering
\includegraphics[width=0.9\textwidth]{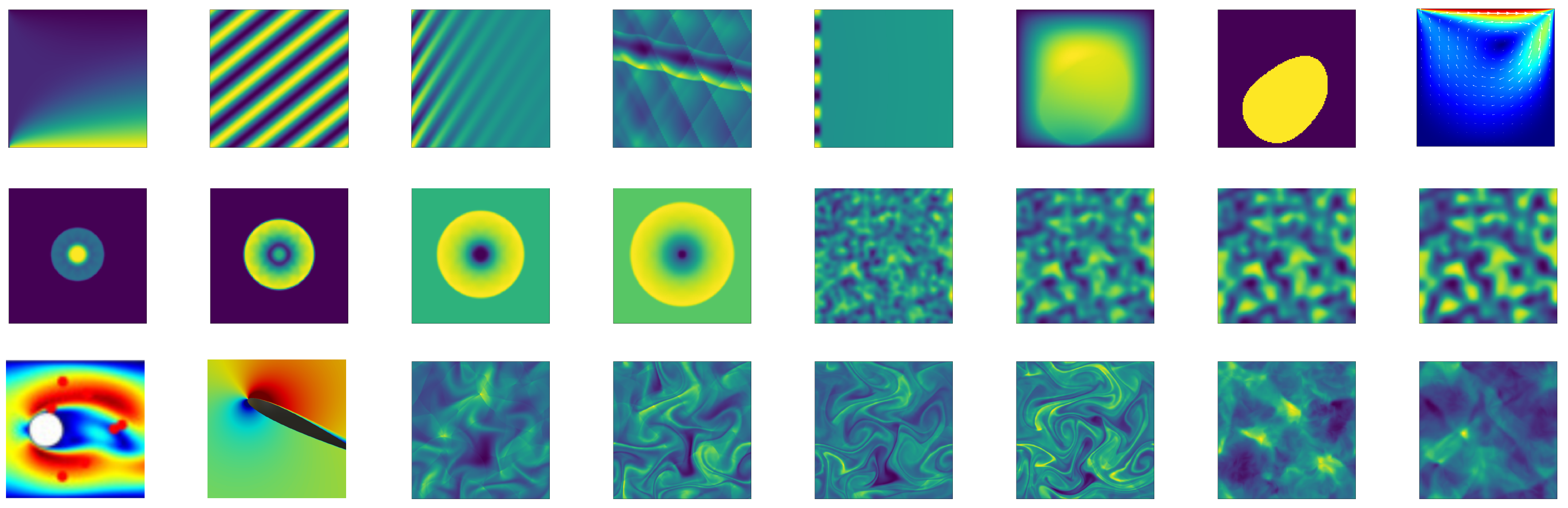}
\vspace{-0.3cm}
\caption{
Demonstration of various datasets of ML for CFD. 
Row 1 from left to right:
(1) 1D Diffusion,
(2)-(3) 1D Advection,
(4) 1D compressible Navier-Stokes,
(5) 1D Reaction-Diffusion,
(6) 2D Darcy flow solution,
(7) 2D Darcy flow coefficient,
(8) Cavity flow.
Row 2 from left to right:
(1)-(4) 2D Shallow Water at 
$t=0.25, 0.5, 0.75, 1$, 
(5)-(8) 2D Reaction-Diffusion at 
$t=1.25, 2.5, 3.75, 5$. 
Row 3 from left to right:
(1) Cylinder flow, 
(2) Airfoil flow, 
(3)-(6) 2D Compressible Navier-Stokes at 
$t=0, 0.5, 1, 2$.,
(7)-(8) 3D compressible Navier-Stokes at $t = 1, 2$.
}
\label{fig:dataset}
\vspace{-0.4cm}
\end{figure*}

\section{Preliminary}
\label{sec:pre}
 
\subsection{Fundamental Theories of Fluid}
Studying fluid problems often involves analyzing the Navier-Stokes (N-S) equations, which describe the motion of fluid substances. The N-S equations are a set of PDEs that describe the motion of viscous fluid substances. These equations are fundamental in the field of fluid dynamics and are key to understanding and predicting the behavior of fluids in various scientific and engineering contexts. Take the incompressible flow as an example, the N-S equation can be simplified as follows:
\begin{equation}
\label{eq:NS}
\left\{
\begin{aligned}
& \nabla \cdot (\rho \mathbf{u}) = 0 \\
& \frac{\partial (\rho \mathbf{u})}{\partial t} + \nabla \cdot (\rho \mathbf{u}\mathbf{u}) = -\nabla p + \nabla \cdot \nu[\nabla \mathbf{u} + (\nabla \mathbf{u})^\top], \\
\end{aligned}
\right.
\end{equation}
where $\rho$ denotes the fluid density, $\nu$ signifies the dynamic viscosity, $\mathbf{u} = (u_x,u_y,u_z)^\top$ is the velocity field, and $p$ is the pressure. 
The first line is the continuity equation, which encapsulates the principle of mass conservation. 
It asserts that the product of density $\rho$ and the velocity field $\mathbf{u}$, when considering a fixed volume, exhibits no net change over time. 
The second line correspond to the momentum equation, articulates the fluid’s momentum variation in reaction to the combined effect of internal and external forces. Sequentially, the terms in momentum equation correspond to the unsteady term, the convective term, the pressure gradient, and the viscous forces, each playing a distinct role in the momentum balance.

In CFD, the governing equations describing fluid motion are often solved numerically due to their inherent complexity and the challenges associated with finding analytical solutions, except in the simplest scenarios. Numerical methods for solving these equations fall into two primary categories: Eulerian and Lagrangian approaches.
\textbf{Eulerian methods} focus on analyzing fluid behavior at fixed locations within the space through which the fluid flows. This approach is particularly effective in fixed grid systems where the grid does not move with the fluid. Eulerian methods are Advectionantageous for problems involving complex boundary interactions and steady-state or quasi-steady flows. They effectively handle multi-phase flow scenarios because they can easily accommodate changes in flow properties at fixed spatial points.
\textbf{Lagrangian methods}, in contrast, track the motion of individual fluid particles as they traverse through the domain. This particle-tracking approach is often more computationally demanding but provides a detailed depiction of the fluid dynamics, making it suitable for capturing unsteady, complex fluid structures and interactions, such as in turbulent flows or when dealing with discrete phases.

\subsection{Traditional Numerical Methods}

In the traditional realm of CFD, numerical methods are employed to discretize the fluid domain into a mesh framework, allowing the transformation of PDEs into solvable algebraic equations. These methods include the finite difference method (FDM)~\cite{taflove2005computational}, which utilizes finite differences to approximate derivatives, effectively transforming the continuous mathematical expressions into discrete counterparts. Another popular approach, the finite volume method (FVM)~\cite{versteeg2007introduction}, involves dividing the domain into control volumes, within which fluxes are calculated to conserve the underlying physical quantities. The finite element method (FEM)~\cite{zienkiewicz2005finite} applies variational principles to minimize error across interlinked elements, offering flexibility in handling complex geometries and boundary conditions. Additionally, the spectral method~\cite{de2000spectral} employs Fourier series to linearize nonlinear problems, proving especially potent under periodic boundary conditions. Lastly, the lattice Boltzmann method (LBM)~\cite{chen1998lattice} adopts a mesoscale perspective, accelerating computations significantly though sometimes at the expense of precision.

\subsection{Benchmark \& Dataset}

Due to the limitations of rapid development of the field, existing benchmarks often cannot comprehensively cover all Advanced ML methods and conduct detailed categorical analyses. Therefore, we summarize existing benchmarks in order to better promote the development of this field. From the famous DeepXDE~\cite{lu2021deepxde} to PDEBench~\cite{takamoto2022pdebench}, and PINNacle~\cite{hao2023pinnacle} to BLASTNet~\cite{chung2024turbulence}, an expanding array of CFD simulation scenarios are being incorporated for comparison. In addition to the widely utilized governing equations that cover a broad spectrum of mathematical properties, the data is also simulated under a variety of settings. These settings include complex geometries, multi-scale phenomena, nonlinear behaviors, and high dimensionality, enriching the diversity and complexity of the simulation scenarios. Furthermore, contributions such as AirFRANS~\cite{bonnet2022airfrans} has made significant strides in addressing specific scenarios like airfoil simulation in aerodynamics more concretely. 

In this context, we provide a thorough review of the prevalent datasets employed to evaluate the performance of ML models in CFD simulations, including 21 PDEs and 13 specific fluid flow problems. PDEs are Advection, Allen-Cahn, Anti-Derivative, Bateman–Burgers, Burgers, Diffusion, Duffing, Eikonal, Elastodynamic, Euler, Gray-Scott, Heat, Korteweg-de Vries, Kuramoto-Sivashinsky, Laplace, Poisson, Reynold Averaged Navier-Stokes, Reaction–Diffusion, Schr\"{o}dinger, Shallow Water, and Wave Equations. Besides, the flow problems are Ahmed-Body, Airfoil, Beltrami flow, Cavity flow, Cylinder flow, Dam flow, Darcy flow, Kovasznay flow, Kolmogorov flow, Navier-Stokes flow, Rayleigh-B\'{e}nard flow, and Transonic flow. Partial demonstration examples are shown in Fig. \ref{fig:dataset}.

\begin{table*}[ht]
\centering
\scriptsize
\caption{Overview of Data-driven Surrogates methods for computational fluid dynamics.}
\vspace{-0.2cm}
\label{tab:method}
\setlength{\tabcolsep}{4.5mm}{
\begin{tabular}{l|c|c|c}
\toprule
\textbf{Methods} & \textbf{Source} & \textbf{Backbone} & \textbf{Scenarios} \\ \hline
\multicolumn{4}{c}{\textbf{\ref{sec:data_driven_surrogates} Data-driven Surrogates}} \\ 
\hline 
\multicolumn{4}{l}{\textbf{\ref{subsubsec:structured_grids} On Regular Grids}} \\ \hline
\rowcolor{lightgray} DPUF~\cite{lee2019data} & \textit{J. Fluid Mech.}  & CNN & CylinderFlow \\ \hline
\rowcolor{lightgray} TF-Net~\cite{wang2020towards}& \textit{KDD 2020}  & CNN & Turbulent flow \\ \hline
\rowcolor{lightgray} EquNet~\cite{wang2020incorporating} &\textit{ICLR 2021}  & CNN & Rayleigh-Bénard, Oceans, Heat \\ \hline
\rowcolor{lightgray} RSteer~\cite{wang2022approximately} &\textit{ICML 2022}  & CNN & smoke \\ \hline
\multicolumn{4}{l}{\textbf{\ref{subsubsec:unstructured_mesh} On Irregular Mesh}} \\ \hline
\rowcolor{lightred} 
 GNS~\cite{sanchez2020learning}& \textit{ICML 2020}  & GNN & Various materials \\ \hline
\rowcolor{lightred} MGN~\cite{pfaff2020learning} & \textit{ICLR 2021}  & GNN & CylinderFlow, Transonic \\ \hline
\rowcolor{lightred} MP-PDE~\cite{brandstetter2022message} & \textit{ICLR 2022}  & GNN & Burgers, Wave, Navier-Stokes \\ \hline
\rowcolor{lightred} Han \textit{\textit{et al.}} \cite{han2022predicting} & \textit{ICLR 2022}  & GNN/Transformer & CylinderFlow, Transonic \\ \hline
\rowcolor{lightred} TIE~\cite{shao2022transformer} &  \textit{ECCV 2022}  & Transformer & FluidFall, FluidShake \\ \hline
\rowcolor{lightred} MAgNet~\cite{boussif2022magnet} &  \textit{NeurIPS 2022}  & GNN/CNN & Burgers \\ \hline
\rowcolor{lightred} GNODE~\cite{bishnoi2022enhancing} &  \textit{ICLR 2023}  & GNN & N-pendulums, Spring  \\ \hline
\rowcolor{lightred} FCN~\cite{he2022flow} & \textit{PHYS FLUIDS.} & GNN & Navier-Stokes  \\ \hline
\rowcolor{lightred} Zhao \textit{et al.}~\cite{zhao2023computationally} & \textit{Arxiv 2023} & GNN & Navier-Stokes  \\ \hline
\rowcolor{lightred} DINo~\cite{yin2022continuous} & \textit{ICLR 2023}  & FourierNet & Wave, Navier-Stokes, Shallow Water \\ \hline
\rowcolor{lightred} LAMP~\cite{wu2023learning} & \textit{ICLR 2023}  & GNN & Nonlinear PDE, Mesh\\ \hline
\rowcolor{lightred} CARE~\cite{luo2023care} & \textit{NeurIPS 2023}  & GNN & CylinderFlow, Particle \\ \hline
\rowcolor{lightred} BENO~\cite{wang2024beno}& \textit{ICLR 2024}  & GNN/Transformer & Poisson \\ \hline
\rowcolor{lightred} HAMLET~\cite{bryutkin2024hamlet}&\textit{Arxiv 2024}  & GNN/Transformer & Shallow Water, Darcy, Diffusion, Airfoil \\ \hline
\multicolumn{4}{l}{\textbf{\ref{subsubsec:lagrangian_particles} On Lagrangian Particles}} \\ \hline
\rowcolor{lightorange} 
 CC~\cite{ummenhofer2019lagrangian} & \textit{ICLR 2020}  & CNN & Particles, DamBreak  \\ \hline
\rowcolor{lightorange} Wessels \textit{et al.}~\cite{wessels2020neural} & \textit{Comput Method Appl M}  & MLP &  DamBreak  \\ \hline
\rowcolor{lightorange} FGN~\cite{li2022graph} & \textit{Comput Graph}  & GNN & Particles  \\ \hline
\rowcolor{lightorange} MCC~\cite{liu2023fast} & \textit{AAAI 2023}  & CNN & DamBreak  \\ \hline 
\rowcolor{lightorange} LFlows~\cite{torres2023lagrangian}& \textit{ICLR 2024}  & Bijective layer & Bird migration  \\ 
\hline 
\rowcolor{lightorange} Li \textit{et al.}~\cite{li2024synthetic}& \textit{Nat. Mach. Intell.}  & Diffusion & 3D Navier-Stokes   \\ \hline
\multicolumn{4}{l}{\textbf{\ref{sec:don} Deep Operator Network}} 
\\ \hline
\rowcolor{lightyellow} DeepONet~\cite{lu2021learning} & \textit{Nat. Mach. Intell.} & MLP & Reaction–Diffusion, other ODEs \\ \hline
\rowcolor{lightyellow} PI-DeepONet~\cite{wang2021learning} & \textit{Sci. Advection} & MLP & Anti-Derivative, Reaction–Diffusion, Burgers, Eikonal, Transonic \\ \hline
\rowcolor{lightyellow} MIONet~\cite{jin2022mionet} & \textit{Siam J Sci Comput} & MLP & ODE, Anti-Derivative, Reaction–Diffusion\\ \hline
\rowcolor{lightyellow} Fourier-MIONet~\cite{jiang2023fourier} & \textit{Arxiv 2023} & MLP & Gas saturation\\ \hline
\rowcolor{lightyellow} NOMAD~\cite{seidman2022nomad} & \textit{NeurIPS 2022} & MLP & Advection, Shallow Water, Anti-Derivative\\ \hline
\rowcolor{lightyellow} Shift-DeepONet~\cite{lanthaler2022nonlinear} & \textit{ICLR 2023} & MLP & Advection, Burgers, Euler \\ 
\hline
\rowcolor{lightyellow} HyperDeepONet~\cite{lee2023hyperdeeponet} & \textit{ICLR 2023} & MLP & Advection, Burgers, Shallow Water \\ 
\hline
\rowcolor{lightyellow} B-DeepONet~\cite{lin2023b} & \textit{J. Comput. Phy.} & MLP & Anti-Derivative, Reaction–Diffusion, Advection  \\ \hline
\rowcolor{lightyellow} SVD-DeepONet~\cite{venturi2023svd} & \textit{Comput Method Appl M} & MLP & ODEs \\ \hline
\rowcolor{lightyellow} L-DeepONet~\cite{kontolati2023learning}& \textit{Arxiv 2023} & MLP & Rayleigh-Bénard, Shallow Water  \\ \hline
\multicolumn{4}{l}{\textbf{\ref{subsubsec:other_operator_learning} In Physical Space}} \\\hline
\rowcolor{lightgreen} MGNO~\cite{li2020multipole}& \textit{NeurIPS 2020} & GNN &  Darcy, Burgers\\ \hline
\rowcolor{lightgreen} Geneva \textit{et al.} \cite{geneva2022transformers} & \textit{Neural Networks} & Transformer &  Navier-Stokes, Lorenz\\ \hline
\rowcolor{lightgreen} GNOT~\cite{hao2023gnot}  & \textit{ICML 2023} & Transformer &  Transonic, Elastodynamic, Navier-Stokes, Darcy\\ \hline
\rowcolor{lightgreen} CNO~\cite{raonic2023convolutional} & \textit{NeurIPS 2023} & CNN &  Poisson, Allen-Cahn, Navier-Stokes, Darcy\\ \hline
\rowcolor{lightgreen} FactFormer~\cite{li2024scalable} &\textit{NeurIPS 2023} &  Transformer  &  Kolmogorov flow, Smoke buoyancy\\ \hline
\rowcolor{lightgreen} LNO~\cite{cao2023lno} & \textit{Arxiv 2023} &  Laplace  &  Diffusion, Duffing, Reaction–Diffusion\\ \hline
\rowcolor{lightgreen} KNO~\cite{xiong2023koopman} & \textit{Arxiv 2023} &  Koopman  &  Bateman–Burgers, Navier-Stokes\\ \hline
\rowcolor{lightgreen} ICON~\cite{yang2023context} &\textit{PNAS} &  Transformer  &  Poisson, Reaction–Diffusion, ODEs\\ \hline
\rowcolor{lightgreen} Transolver~\cite{wu2024transolver} &\textit{ICML 2024} &  Transformer  &  CarDesign, Airfoil\\ \hline
\multicolumn{4}{l}{\textbf{\ref{subsubsec:f_operator_learning} In Fourier Space}} 
\\ \hline
\rowcolor{lightblue} FNO~\cite{li2020fourier} & \textit{ICLR 2021} & Fourier &  Burgers, Darcy, Navier-Stokes \\ \hline
\rowcolor{lightblue} PINO~\cite{li2021physics} & \textit{ACM/IMS Trans. Data Sci.} & Fourier &  Burgers, Darcy, Navier-Stokes, Kolmogorov  \\ \hline
\rowcolor{lightblue} GEO-FNO~\cite{li2022fourier} & \textit{JMLR} & Fourier &  Advection, Elastodynamic, Airfoil, Navier-Stokes \\ \hline
\rowcolor{lightblue} U-NO~\cite{rahman2022u} & \textit{TMLR} & Fourier &  Darcy, Navier-Stokes \\ \hline
\rowcolor{lightblue} SSNO~\cite{rafiq2022ssno} & \textit{IEEE Access} & Fourier &  Burgers, Darcy, Navier-Stokes \\ \hline
\rowcolor{lightblue} F-FNO~\cite{tran2022factorized} & \textit{ICLR 2023} & Fourier &  Kolmogorov, Transonic  \\ \hline
\rowcolor{lightblue} CFNO~\cite{brandstetter2022clifford} & \textit{ICLR 2023} & Fourier &  Navier-Stokes, Shallow Water, Maxwell  \\ \hline
\rowcolor{lightblue} CMWNO~\cite{xiao2022coupled}& \textit{ICLR 2023} & Fourier & Gray-Scott   \\ \hline
\rowcolor{lightblue} MCNP~\cite{zhang2023monte} & \textit{Arxiv 2023} & Fourier &  Diffusion, Allen-Cahn, Navier-Stokes  \\ \hline
\rowcolor{lightblue} $G$-FNO~\cite{helwig2023group} & \textit{ICML 2023} & Fourier &  Navier-Stokes, Shallow Water\\ \hline
\rowcolor{lightblue} GINO~\cite{li2023geometry} & \textit{NeurIPS 2023} & Fourier &  Ahmed-Body, Reynold Averaged Navier-Stokes\\ \hline
\rowcolor{lightblue} DAFNO~\cite{liu2023domain} & \textit{NeurIPS 2023} & Fourier &  Hyperelasticity, Airfoil\\ \hline
\rowcolor{lightblue} FCG-NO~\cite{rudikov2024neural}  & \textit{ICML 2024} &  Fourier  &  Poisson, Diffusion\\ \hline
\bottomrule
\end{tabular}}
\vspace{-0.3cm}
\end{table*}

\section{Data-driven Surrogates}
\label{sec:data_driven_surrogates}

Data-driven Surrogates are models that rely solely on observed data to train algorithms capable of simulating complex fluid dynamics, which have experienced swift advancements. These models are impactful and can be broadly classified based on their approach to spatial discretization, ranging from methods that: \textbf{1) Dependent on discretization}, \textbf{2) Independent of discretization}. The former requires dividing the data domain into a specific grid, mesh or particle structure and designing the model architecture, while the latter does not rely on discretization techniques, but instead directly learns the solution in the continuous space.

\subsection{Dependent on Discretization}
\label{subsec:dependent_discretization}

We categorize these methods based on the type of discretization into three categories: 1) \underline{\textbf{on Regular grid}}, 2) \underline{\textbf{on Irregular mesh}}, 
and 3) \underline{\textbf{on Lagrangian particles}}.

\subsubsection{Regular Grids}
\label{subsubsec:structured_grids}
This section provides a comprehensive review of pivotal contributions in regular grid approaches. These innovative methods have established a solid groundwork for harmonizing neural network architectures, particularly CNNs, with CFD for enhanced predictive capabilities, marking them as trailblazers in the field.
DPUF~\cite{lee2019data} is the first to implement CNNs on regular grids for learning transient fluid dynamics. Their approach is notable for incorporating physical loss functions, which explicitly enforces the conservation of mass and momentum within the networks. This integration of physical principles and ML marks a significant step in the field.
And then TF-Net~\cite{wang2020towards} aims to predict turbulent flow by learning from the highly nonlinear dynamics of spatiotemporal velocity fields. These fields are derived from large-scale fluid flow simulations, pivotal in turbulence and climate modeling. This work represents a significant effort in understanding complex fluid behaviors through ML.
Incorporating theoretical principles into ML, EquNet~\cite{wang2020incorporating}, seeks to improve accuracy and generalization in fluid dynamics modeling. It achieves this by incorporating various symmetries into the learning process, thereby enhancing the model's robustness and theoretical grounding. RSteer~\cite{wang2022approximately} presentes a novel class of approximately equivariant networks. These networks were specifically designed for modeling dynamics that are imperfectly symmetric, relaxing equivariance constraints. This approach offered a new perspective on handling complex dynamic systems in fluid dynamics. Collectively, these studies demonstrate the burgeoning potential of ML in fluid dynamics. They underscore the effectiveness of combining traditional fluid dynamics principles with data-driven techniques, particularly in the realms of regular grids and discretization-dependent methods.

\subsubsection{Irregular Mesh}
\label{subsubsec:unstructured_mesh}


Irregular mesh challenges regular grid-based surrogates, prompting the adoption of adaptable solutions like Graph Neural Networks (GNNs) for diverse structures and sizes. The GNS model~\cite{sanchez2020learning} and MeshGraphNets~\cite{pfaff2020learning} both employ GNNs, with the former representing physical systems as particles in a graph using learned message-passing for dynamics computation, and the latter simulating complex phenomena on irregular meshes, demonstrating the networks' proficiency in handling irregular topology. Further enhancing these approaches, MP-PDE~\cite{brandstetter2022message} replaces heuristic components with backprop-optimized neural function mapping, refining PDE solutions, while LAMP~\cite{wu2023learning} optimizes spatial resolutions to focus resources on dynamic regions. Additionally, the GNODE model~\cite{bishnoi2022enhancing} introduces a graph-based neural ordinary differential equation to learn the time evolution of dynamical systems, and the Flow Completion Network (FCN)\cite{he2022flow} uses GNNs to infer fluid dynamics from incomplete data. Recognizing the limitations of CNNs, Zhao \textit{et al.}\cite{zhao2023computationally} propose a novel model that integrates CNNs with GNNs to better capture connectivity and flow paths within porous media. Addressing the challenge of training GNNs on high-resolution meshes, MS-GNN-Grid~\cite{lino2022multi}, MS-MGN~\cite{fortunato2022multiscale}, and BSMS-GNN~\cite{cao2023efficient} explore strategies to manage mesh resolution and connectivity, with the latter providing a systematic solution without the need for manual mesh drawing. Complementing these developments, CARE~\cite{luo2023care} utilizes a context variable from trajectories to model dynamic environments, enhancing adaptability and precision.
BENO~\cite{wang2024beno} introduces a boundary-embedded neural operator based on graph message passing to incorporate complex boundary shape into PDE solving. It is similarly employed in HAMLET~\cite{bryutkin2024hamlet}, which utilizes graph transformers alongside modular input encoders to seamlessly integrate PDE information into the solution process.

GNN is naturally suitable for irregular mesh, however, they are either steady-state or next-step prediction models, which often experience drift and accumulate errors over time. In contrast, sequence models leverage their extended temporal range to identify and correct drifts. Han \textit{et al.} \cite{han2022predicting} first uses GNN to aggregate local data, then coarsens the graph to pivotal nodes in a low-dimensional latent space. A transformer model predicts the next latent state by focusing on the sequence, and another GNN restores the full graph.
Additionally, TIE~\cite{shao2022transformer} effectively captures the complex semantics of particle interactions without using edges, by modifying the self-attention module to mimic the update mechanism of graph edges in GNNs.

Another technical approach is based on coordinate-based Implicit Neural Representation (INR) networks, which allow the model to operate without the need for interpolating or regularizing the input grid. 
For example, MAgNet~\cite{boussif2022magnet} employs INR to facilitate zero-shot generalization to new non-uniform meshes and to extend predictions over longer time spans. Besides, DINo~\cite{yin2022continuous} models the flow of a PDE using continuous-time dynamics for spatially continuous functions, enabling flexible extrapolation at arbitrary spatio-temporal locations.

\subsubsection{Lagrangian Particles}
\label{subsubsec:lagrangian_particles}

Fluid representations vary, with particle-based Lagrangian representations being also popular due to their wide usage. Yet, fluid samples typically comprise tens of thousands of particles, especially in complex scenes.
CC~\cite{ummenhofer2019lagrangian} utilizes a unique type of convolutional network, which acts as the main differentiable operation, effectively linking particles to their neighbors and simulating Lagrangian fluid dynamics with enhanced precision.
Then, Wessels \textit{et al.}~\cite{wessels2020neural} combines PINNs with the updated Lagrangian method to solve incompressible free surface flow problems.
FGN~\cite{li2022graph} conceptualizes fluid particles as nodes within a graph, with their interactions depicted as edges.
Prantl \textit{et al.}~\cite{prantl2022guaranteed} designs architecture to conserve momentum.
Furthermore, Micelle-CConv~\cite{liu2023fast} introduces a dynamic multi-scale gridding method. It aims to minimize the number of elements that need processing by identifying and leveraging repeated particle motion patterns within consistent regions.
More recently, LFlows~\cite{torres2023lagrangian} models fluid densities and velocities continuously in space and time by expressing solutions to the continuity equation as time-dependent density transformations through differentiable and invertible maps.
Li \textit{et al.}~\cite{li2024synthetic} use a diffusion model to accurately reproduce the statistical and topological properties of particle trajectories in Lagrangian turbulence.

\subsection{Independent of Discretization}
\label{subsec:independent_discretization}

Neural Operator is a powerful means to achieve the goal of being independent of discretization. It learns the solution mapping between two infinite-dimensional function spaces for PDE solving. 
Let $\mathcal{N}$ be a neural operator that aims to approximate the solution operator $\mathcal{S}$ of the fluid governing equations (i.e., NS Equation). For given fluid dynamics parameters $\mu$ (i.e., Reynolds number), we seek the velocity field $v$ and pressure field $p$. It is formalized as follows:

\begin{equation}
\mathcal{N}: \mathcal{M} \to \mathcal{H},
\end{equation}
\begin{equation}
\mu \mapsto (v, p) \approx \mathcal{S}[\mu],
\end{equation}
where $\mathcal{M}$ is the parameter space, $\mathcal{H}$ is the function space composed of velocity and pressure fields, $\mathcal{S}$ is the exact solution operator of the NS equations, and $\mathcal{N}[\mu]$ is the approximate solution given by the neural operator.
We categorize existing methods by the means of realizing the integral function approximation. 1)\underline{\textbf{ Deep Operator Network.}} It approximates operators by dividing the input into two branches: one branch learns representations of the function, while the other captures the specific points at which the function is evaluated. 
2)\underline{\textbf{ In Physical Space.}} It leverages the flexibility of neural networks on physical spaces (i.e., GNNs, CNNs) to model complex relationships in data.
3)\underline{\textbf{ In Fourier Space.}} It utilizes the Fourier transform to model the integral operators in a spectral space to capture global dependencies.

\subsubsection{Deep Operator Network}
\label{sec:don}

DeepONet~\cite{lu2021learning} represents a significant advance in neural operator theory, marking the transition towards learning mappings between functional spaces. 
Following this pioneering work, PI-DeepONet~\cite{wang2021learning} uses PDE residuals for unsupervised training, enhancing the ability to learn without explicit supervision. MIONet~\cite{jin2022mionet} and Fourier-MIONet~\cite{jiang2023fourier} expand on this by introducing a neural operator with branch nets for input function encoding and a trunk net for output domain encoding, with the latter incorporating the FNO to model multi-phase flow dynamics under varied conditions. NOMAD~\cite{seidman2022nomad} further innovates by introducing a nonlinear decoder map that can model nonlinear submanifolds within function spaces. HyperDeepONet~\cite{lee2023hyperdeeponet} and Shift-DeepONet~\cite{lanthaler2022nonlinear} respectively utilize a hyper-network to reduce parameter count while enhancing learning efficiency, and a sophisticated nonlinear reconstruction mechanism to approximate discontinuous PDE solutions. 
Additionally, B-DeepONet~\cite{lin2023b} incorporates a Bayesian framework using replica exchange Langevin diffusion to optimize training convergence and uncertainty estimation. 
SVD-DeepONet~\cite{venturi2023svd} and L-DeepONet~\cite{kontolati2023learning} employ methods derived from proper orthogonal decomposition and latent representations identified by auto-encoders to improve model design and handle high-dimensional PDE functions, respectively.

\subsubsection{In Physical Space}
\label{subsubsec:other_operator_learning}
Implementing functional mapping in physical spaces using diverse network architectures has led to the development of novel neural operators.
MGNO~\cite{li2020multipole} utilizes a class of integral operators, with the kernel integration being computed through graph-based message passing on different GNNs.
Besides, LNO~\cite{cao2023lno} enhances interpretability and generalization ability, setting it apart from Fourier-based approach by leveraging a more intrinsic mathematical relationship in function mappings.
GNOT~\cite{hao2023gnot} introduces a scalable and efficient transformer-based framework featuring heterogeneous normalized attention layers, which offers exceptional flexibility to accommodate multiple input functions and irregular meshes.
Recently, CNO~\cite{raonic2023convolutional} adapts CNNs to handle functions as inputs and outputs. CNO offers a unique approach that maintains continuity even in a discretized computational environment, diverging from FNO’s emphasis on Fourier space to focus on convolutional processing.
Furthermore, Geneva \textit{et al.} \cite{geneva2022transformers} and KNO~\cite{xiong2023koopman} center on approximating the Koopman operator. It acts on the flow mapping of dynamic systems, enabling the solution of an entire family of non-linear PDEs through simpler linear prediction problems.

Additionally, Transformer architecture is also applied.
FactFormer~\cite{li2024scalable} introduces a low-rank structure that enhances model efficiency through multidimensional factorized attention.
Besides, In-Context Operator Networks (ICON)~\cite{yang2023context} revolutionizes operator learning by training a neural network capable of adapting to different problems without retraining, contrasting with methods that require specific solutions or retraining for new problems. More recently, Transolver~\cite{wu2024transolver} has introduced a novel physics-based attention mechanism, which adaptively divides the discretized domain to effectively capture complex physical correlations.

\subsubsection{In Fourier Space}
\label{subsubsec:f_operator_learning}

The Fourier Neural Operator (FNO)~\cite{li2020fourier} marks a significant development in neural operators by innovatively parameterizing the integral kernel directly in Fourier space, thus creating an expressive and efficient architecture. Expanding on this concept, GEO-FNO~\cite{li2022fourier} and GINO~\cite{li2023geometry} address the complexities of solving PDEs on arbitrary geometries and large-scale variable geometries, respectively. U-NO~\cite{rahman2022u} enhances the structure with a U-shaped, memory-enhanced design for deeper neural operators, while PINO~\cite{li2021physics} combines training data with physics constraints to learn solution operators even without training data.
Further advancements include F-FNO~\cite{tran2022factorized}, which incorporates separable spectral layers, augments residual connections, and applies sophisticated training strategies. Similarly, Rafiq \textit{et al.}~\cite{rafiq2022dsfa} utilizes spectral feature aggregation within a deep Fourier neural network. SSNO~\cite{rafiq2022ssno} integrates spectral and spatial feature learning, and MCNP~\cite{zhang2023monte} progresses unsupervised neural solvers with probabilistic representations of PDEs. 
DAFNO~\cite{liu2023domain} and RecFNO~\cite{zhao2024recfno} respectively address surrogate models for irregular geometries and evolving domains, and improve accuracy and mesh transferability. CoNO~\cite{tiwari2023cono} delves deeper by parameterizing the integral kernel within the complex fractional Fourier domain. Choubineh \textit{et al.}~\cite{choubineh2023fourier} and CMWNO~\cite{xiao2022coupled} use FEM-calculated outputs as benchmarks and decouple integral kernels during multiwavelet decomposition in Wavelet space, enhancing the model's analytical capabilities.
CFNO~\cite{brandstetter2022clifford} integrates multi-vector fields with Clifford convolutions and Fourier transforms, addressing time evolution in correlated fields. $G$-FNO~\cite{helwig2023group} innovates by extending group convolutions into the Fourier domain, crafting layers that maintain equivariance to spatial transformations like rotations and reflections, thus enhancing model versatility.

\begin{table*}[ht]
\scriptsize
\centering
\caption{Overview of Physics-driven Surrogates methods for computational fluid dynamics.}
\vspace{-0.2cm}
\label{tab:method_physics}
\setlength{\tabcolsep}{2.5mm}{
\begin{tabular}{l|c|c|c}
\toprule
\textbf{Methods} & \textbf{Source} & \textbf{Backbone} & \textbf{Scenarios} \\ \hline
\multicolumn{4}{c}{\textbf{\ref{sec:physics_driven_surrogates} Physics-driven Surrogates}} \\ 
\hline
\multicolumn{4}{l}{\textbf{\ref{subsec:pinn} Physics-Informed Neural Networks}} \\ \hline
\rowcolor{lightcyan} PINN~\cite{raissi2019physics} & \textit{J. Comput. Phy.}  & MLP & Allen-Cahn, Navier-Stokes, Schrodinger, Korteweg-de Vries, Burgers \\ \hline
\rowcolor{lightcyan} Bai et al.~\cite{bai2020ApplyingPhysicsInformed} & \textit{J. Hydrodyn.} &  MLP & Propagating wave, Lid-driven cavity flow, Turbulent flows \\ \hline
\rowcolor{lightcyan} PINN-SR~\cite{chen2021physics} & \textit{Nat. Commun.} & MLP & Kuramoto-Sivashinsky, Navier-Stokes, Schrodinger, Reaction–Diffusion, Burgers \\ \hline
\rowcolor{lightcyan} NSFnet~\cite{jin_nsfnets_2021} & \textit{J. Comput. Phys.} & MLP & Kovasznay, CylinderFlow, 3D Beltrami \\ \hline
\rowcolor{lightcyan} Phygeonet~\cite{gao2021phygeonet} & \textit{J. Comput. Phys.} & CNN & Heat, Poisson, Navier-Stokes \\ \hline
\rowcolor{lightcyan} PINN-LS~\cite{mowlavi2021optimal} & \textit{APS DFD Meet. Abstr.} & MLP & Laplace, Burgers, Kuramoto-Sivashinsky, Navier-Stokes \\ \hline

\rowcolor{lightcyan} ResPINN~\cite{cheng2021deep} & \textit{Water} & MLP & Burgers, Navier-Stokes \\ \hline
\rowcolor{lightcyan} CPINN~\cite{zeng2022competitive} & \textit{Arxiv 2022} & MLP &  Poisson, Schrodinger, Burgers, Allen-Cahn \\ \hline
\rowcolor{lightcyan} stan~\cite{gnanasambandam2022self} & \textit{Arxiv 2022} & MLP & Dirichlet, Neumann, Klein-Gordon, Heat\\ \hline
\rowcolor{lightcyan} Meta-Auto-Decoder~\cite{huang2022meta} & \textit{NeurIPS 2022} & MLP & Burgers, Laplace, Maxwell \\ \hline
\rowcolor{lightcyan} BINN~\cite{sun2023binn} & \textit{Comput Method Appl M} & MLP & Poisson, Navier-Stokes \\ \hline
\rowcolor{lightcyan} Hodd-PINN ~\cite{you2023high} & \textit{Arxiv 2023} & Resnet & Convection, Heat \\ \hline

\rowcolor{lightcyan} DATS~\cite{toloubidokhti2023dats} & \textit{ICLR 2024} & MLP & Burgers, Reaction–Diffusion, Helmholtz, Navier-Stokes \\ \hline
\rowcolor{lightcyan} PINNsFormer~\cite{zhao2023pinnsformer} & \textit{ICLR 2024} & MLP/Transformer & Reaction–Diffusion, Wave, Navier-Stokes \\ \hline
\rowcolor{lightcyan} PeRCNN~\cite{rao2023encoding} & \textit{Nat. Mach. Intell.} & CNN & Reaction–Diffusion, Burgers, Kolmogorov \\ \hline
\rowcolor{lightcyan} NASPINN~\cite{wang2024pinn} & \textit{J. Comput. Phys.} & MLP & Burgers, Advection, Poisson \\ \hline
\multicolumn{4}{l}{\textbf{\ref{subsec:discritized_pde} Discretized Constraint-Informed Neural Network}} \\ \hline
\rowcolor{lightpurple} DGM~\cite{sirignano2018dgm} & \textit{J. Comput. Phys.} & RNN & Free Boundary PDE  \\ \hline
\rowcolor{lightpurple} AmorFEA~\cite{xue2020amortized} & \textit{ICML 2020} & MLP & Poisson  \\ \hline
\rowcolor{lightpurple} EDNN~\cite{du2021evolutional} & \textit{Phys. Rev. E} & MLP & Advection, Burgers, Navier-Stokes, Kuramoto-Sivashinsky  \\ \hline
\rowcolor{lightpurple} Liu \textit{et al.}\cite{liu2022predicting}& \textit{Commun. Phys.} & CNN & Reaction–Diffusion, Burgers, Navier-Stokes \\ \hline
\rowcolor{lightpurple} Gao \textit{et al.}\cite{gao2022physics}& \textit{Comput Method Appl M} & GNN & Poisson, Elastodynamic, Navier-Stokes \\ \hline
\rowcolor{lightpurple} ADLGM~\cite{aristotelous2023adlgm} & \textit{J. Comput. Phys.} & MLP & Poisson, Burgers, Allen-Cahn  \\ \hline
\rowcolor{lightpurple} INSR~\cite{chen2023implicit} & \textit{ICML 2023} & MLP & Advection, Euler, Elastodynamic  \\ \hline
\rowcolor{lightpurple} Neu-Gal~\cite{bruna2024neural} & \textit{J. Comput. Phys.} & MLP & Korteweg-de Vries, Allen-Cahn, Advection  \\ \hline
\rowcolor{lightpurple} PPNN~\cite{liu2024multi} & \textit{Commun. Phys.} & CNN & Reaction–Diffusion, Burgers, Navier-Stokes  \\ \hline
\rowcolor{lightpurple} FEX~\cite{song2024finite} & \textit{Arxiv 2024} & Tree & Hindmarsh-Rose, FHN \\ \hline
\bottomrule
\end{tabular}}
\vspace{-0.3cm}
\end{table*}

\section{Physics-driven Surrogates}
\label{sec:physics_driven_surrogates}

Although data-driven models have demonstrated potential in CFD simulations, they are not without their challenges, such as the significant expense associated with data collection and concerns over their generalization and robustness. Consequently, integrating physics-based priors is crucial, leveraging the power of physical laws to enhance model reliability and applicability. We categorize them based on the type of embedded knowledge into: \underline{\textbf{1) Physics-Informed, 2) Constraint-Informed}}. The former transforms physical knowledge into constraints for the neural network, ensuring that predictions adhere to known physical principles. The latter draws inspiration from traditional PDE solvers, integrating these approaches into the neural network's training process.

\subsection{Physics-Informed Neural Network (PINN)}
\label{subsec:pinn}
The development of PINN marks a significant evolution, blending deep learning with physical laws to solve complex fluid governing differential equations. 
Formally, PINN utilizes the neural network that approximates a function $u(x)$, subject to a differential equation $\mathcal{D}(u) = 0$ over a domain $\Omega$ and boundary conditions $\mathcal{B}(u) = 0$ on the boundary $\partial \Omega$. The essence of PINN is captured by the loss function:
\begin{equation}
    \mathcal{L}(\theta) = \mathcal{L}_{data}(\theta) + \mathcal{L}_{physics}(\theta),
\end{equation}

where $\mathcal{L}_{data}$ is the data-driven term ensuring fidelity to known solutions or measurements, and $\mathcal{L}_{physics}$ encodes the physical laws, typically differential equations, governing the system. This is mathematically represented as:
\begin{equation}
    \mathcal{L}_{data}(\theta) = \frac{1}{N} \sum_{i=1}^{N} \| u_{\theta}(x_i) - u(x_i) \|^2,
\end{equation}
\begin{equation}
    \mathcal{L}_{physics}(\theta) = \frac{1}{M} \sum_{j=1}^{M} \| \mathcal{D}(u_{\theta})(x_j) \|^2 + \frac{1}{P} \sum_{k=1}^{P} \| \mathcal{B}(u_{\theta})(x_k) \|^2,
\end{equation}
with $\theta$ denoting the parameters of the neural network, $u_{\theta}$ the neural network's approximation of $u$, and $x_i$, $x_j$, $x_k$ sampled points from the domain and boundary.

Building upon this foundational concept, the seminal work~\cite{raissi2019physics} lays the foundation for PINN, demonstrating their capability in solving forward and inverse problems governed by differential equations. Subsequent advancements, PINN-SR~\cite{chen2021physics} further enhances PINN by integrating deep neural networks for enriched representation learning and sparse regression, thereby refining the approximation of system variables. NSFnet~\cite{jin_nsfnets_2021} introduces a breakthrough with Navier-Stokes flow nets (NSFnets), specializing PINN for simulating incompressible laminar and turbulent flows by directly encoding governing equations, thus reducing the reliance on labeled data. 
Advancing into more sophisticated domains, Meta-Auto-Decoder~\cite{huang2022meta} leverages a mesh-free and unsupervised approach, utilizing meta-learning to encode PDE parameters as latent vectors. This innovation allows for quick adaptation of pre-trained models to specific equation instances, enhancing flexibility and efficiency. Bai et al. \cite{bai2020ApplyingPhysicsInformed} expand PINNs' applications in fluid dynamics, particularly in simulating flow past cylinders without labeled data, by transforming equations into continuum and constitutive formulations, showcasing PINNs' potential in flow data assimilation. Similarly, Phygeonet \cite{gao2021phygeonet} has developed a unique approach by morphing between complex meshes and a square domain and proposing a novel physics-constrained CNN architecture that enables learning on irregular domains without relying on labeled data.


Technological enhancements in the field include the development of PINN-LS \cite{mowlavi2021optimal}, which optimizes the learning process by treating the objective function as a regularization term with adaptive weights. The integration of ResNet blocks into PINNs, termed ResPINN \cite{cheng2021deep}, has been crucial for solving fluid flows dependent on partial differential equations and enables precise predictions of velocity and pressure fields across spatio-temporal domains. Moreover, competitive PINNs like CPINN \cite{zeng2022competitive} enhance model accuracy by training a discriminator to identify and correct errors, and Stan et al. \cite{gnanasambandam2022self} have introduced the smooth Stan function to streamline the gradient flow necessary for computing derivatives and scaling the input-output mapping, improving learning efficiency. 
There also have been notable integrations and applications of PINNs. For instance, Binn \cite{sun2023binn} has successfully merged PINNs with the boundary integral method, facilitating their use in complex geometrical scenarios. Additionally, the integration of high-order numerical schemes into PINNs is exemplified by Hodd-PINN \cite{you2023high}, which combines high-order finite difference methods, Weighted Essentially Non-Oscillatory (WENO) discontinuity detection, and traditional PINNs to bolster their capability in modeling complex fluid dynamics.

However, traditional meta-learning approaches often treat all Physics Informed Neural Network (PINN) tasks uniformly. To address this, DATS \cite{toloubidokhti2023dats} advances the field by deriving an optimal analytical solution that tailors the sampling probability of individual PINN tasks to minimize their validation loss across various scenarios. 

Furthermore, PeRCNN \cite{rao2023encoding} introduces a novel physically encoded architecture that embeds prior physical knowledge into its structure, leveraging a spatio-temporal learning paradigm to aim for a robust, universal model with enhanced interpretability and adherence to physical principles. Additionally, the PINNsFormer \cite{zhao2023pinnsformer} brings architectural innovation with transformers, accurately approximating PDE solutions by leveraging multi-head attention mechanisms to effectively capture temporal dependencies. More recently, NAS-PINN \cite{wang2024pinn} propels the field forward by introducing a neural architecture search-guided method for PINNs, automating the search for optimal neural architectures tailored to solve specific PDEs, pushing the boundaries of what can be achieved with ML in physical sciences.

\subsection{Discretized Constraint-Informed Neural Network}
\label{subsec:discritized_pde}

Recent studies have explored to merge the core principles of PDE equations with neural network architectures to address complex fluid dynamics problems, which we call discretized Constraint-informed neural networks. 
Representative of this approach is the Deep Galerkin Method (DGM)~\cite{sirignano2018dgm}, which approximates high-dimensional PDE solutions using deep neural networks trained to satisfy differential operators, initial conditions, and boundary conditions. Similarly, EDNN~\cite{du2021evolutional} numerically updates neural networks to predict extensive state-space trajectories, enhancing parameter space navigation. AmorFEA~\cite{xue2020amortized} combines the accuracy of PDE solutions with the advantages of traditional finite element methods. Additionally, Liu \textit{et al.}~\cite{liu2022predicting} incorporate partially known PDE operators into the CNN kernel, improving stability for extended roll-outs, while Gao \textit{et al.}~\cite{gao2022physics} integrate PINNs with adaptive mesh using the Galerkin method to reduce training complexity on general geometries.
Further developments include INSR~\cite{chen2023implicit}, which integrates classical time integrator into neural networks to effectively address non-linearity. ADLGM~\cite{aristotelous2023adlgm} and NeuGal~\cite{bruna2024neural} refine this integration through adaptive sampling and Neural Galerkin schemes, respectively, both enhancing learning of PDE solutions through active learning strategies. Moreover, PPNN~\cite{liu2024multi} innovatively embeds physics-based priors into its architecture by mapping discretized governing equations to network structures, highlighting the deep connections between PDE operators and network design.
FEX~\cite{song2024finite} represents dynamics on complex networks using binary trees composed of finite mathematical operators with minor prior knowledge on complex networks.


\section{ML-assisted Numerical Solutions}
Despite advancements in end-to-end surrogate modeling, they have yet to match the accuracy of numerical solvers, especially for long-term rollouts where error accumulation becomes significant, and in scenarios involving unseen working conditions during training. Consequently, researchers are exploring a blend of ML and numerical solvers, carefully replacing only parts of the numerical solver to balance speed, accuracy, and generalization. We categorize these methods into three primary classes: \textbf{1) enabling accurate simulations at coarser resolutions or with fewer degrees of freedom, ranging from learning discretization schemes and fluxes, closure modeling, and reduced modeling}; \textbf{2) employing learned preconditioners to accelerate linear system solutions}; and \textbf{3) a range of miscellaneous techniques, ranging from super-resolution to correcting iterative steps.}
\label{sec:ml_assisted_solutions}

\subsection{Assist Simulation at Coarser Scales}
\label{subsec:discretization_scheme}
In numerical methods such as finite-difference (FD) and finite-volume (FV), discretization errors arise as mesh resolution coarsens. The fundamental idea behind learnable discretization schemes or fluxes is to generate space- and time-varying FD or FV coefficients, or corrections to standard coefficients, that maintain high accuracy even at coarse mesh resolutions. In the study by \cite{bar2019learning}, FV coefficients are learned and applied to solve 1D equations such as Burgers, Kuramoto-Sivashinsky, and Korteweg-de Vries. Building on this, Kochkov \textit{et al.} \cite{kochkov2021machine} and List \textit{et al.} \cite{list2022learned} use CNNs to learn fluxes, extending the approach to 2D turbulent flows. \cite{despres2020machine} introduces learnable fluxes for bi-material 2D compressible flows with complex boundary shapes at coarse resolutions. This method involves learning fluxes for various geometric primitives and approximating general shapes by decomposing them into these primitives.

Regarding simulations with larger time steps, Zhang \textit{et al.} \cite{zhang2022multiscale} focus on learning to decouple the stiff chemical reaction component from the coupled flow-combustion system, allowing the flow to be simulated by any time steps. More recently, Sun \textit{et al.} \cite{sun2023neural} tackle the issue of substantial information loss in down-sampled temporal resolutions. They combined time-series sequencing and accounted for more than just the most recent states to predict temporal terms with enriched information.

\subsection{Preconditioning}
\label{subsec:multi_grid}
In the simulation of incompressible flows, the fractional step method (also known as the advection-projection method) has proven to be highly effective and is increasingly attracting attention in the field. This method involves two primary stages: firstly, the prediction step, which advects the current velocity $u_n$ to an intermediate state $u_{n+1}^*$, ignoring the influence of the pressure gradient; and secondly, the pressure-projection step, which ensures the incompressibility constraint, $\nabla \cdot u_{n+1} = 0$, by projecting $u_{n+1}^*$ into a divergence-free space. The updated velocity is determined according to Eq.~\ref{eq:pressure:correction}:

\begin{equation}
\label{eq:pressure:correction}
u_{n+1} = u_{n+1}^* - (\Delta t/\rho)\nabla p_{n+1},
\end{equation}
where the new pressure field is governed by Eq.~\ref{eq:pressure:projection}:
\begin{equation}
\label{eq:pressure:projection}
\nabla^2 p_{n+1} = \frac{\rho}{\Delta t} \nabla \cdot u_{n+1}^*.
\end{equation}

Solving Eq.~\ref{eq:pressure:projection} often involves the preconditioned conjugate gradient (PCG) method, where traditionally, the design of preconditioners necessitates expert domain knowledge. However, advancements in ML have begun to provide a data-driven approach for developing preconditioners that surpass traditional methods in performance. Sappl \textit{et al.} \cite{sappl2019deep} employ CNNs to learn preconditioners in regular domains, demonstrating superior performance over traditional methods.
In the context of improving MultiGrid preconditioners, Greenfeld \textit{et al.} \cite{greenfeld2019learning} develop a model that maps local discretization information to local prolongation matrices, relying the spectral radius of the error propagation matrix as a self-supervised loss. This approach does not require labeled data, and outperforms previous supervised, black-box models.
Subsequently, Luz \textit{et al.} \cite{luz2020learning} adopt a similar unsupervised fashion but extends learning prolongation operators to complex meshes, utilizing GNNs as the backbone.

\subsection{Miscellaneous}
\label{subsec:adhoc_combine}
Besides the methods above, there exist other ML-assisted approaches that are harder to categorize. Pathak \textit{et al.} \cite{pathak2020using} and Um \textit{et al.} \cite{um2020solver} both implement a strategy where they time-stepped PDEs at a coarser scale and then appended ML modules to upscale the results, aiming to align them with high-resolution simulation results. Their work primarily focused on regular domains. In a similar path, CFD-GCN~\cite{belbute2020combining} utilizes Graph Convolutional Networks (GCNs) to conduct upscaling, allowing for this application to the general mesh.
CFDNet~\cite{obiols2020cfdnet} focuses on accelerating the convergence with fewer iterative steps by learning to correct the temporary solutions during iterations.

\section{Inverse Design \& Control}
\label{sec:lab}

\begin{figure}[t]
\centering
\includegraphics[width=0.48\textwidth]{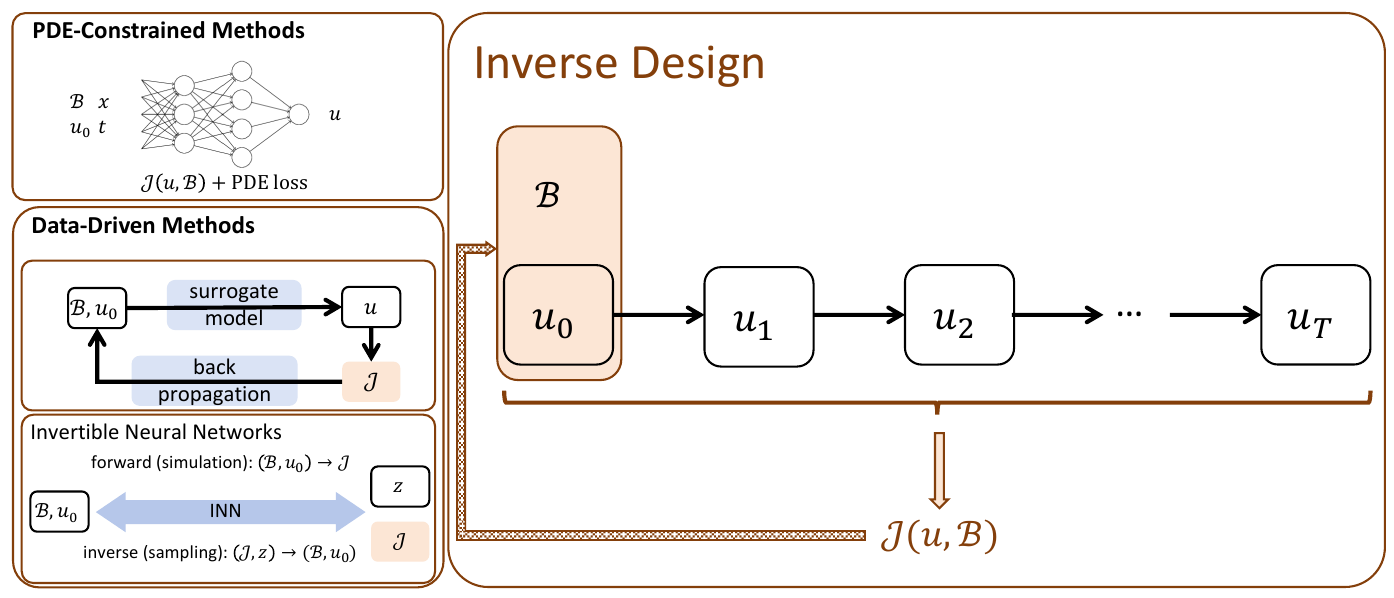}
\vspace{-0.1cm}
\caption{Demonstration of inverse design to optimize the design parameters. We review existing methods with a novel classification including PDE-constrained Methods and Data-driven Methods.}
\label{fig:inverse_design}
\vspace{-0.4cm}
\end{figure}

\subsection{Inverse Design}
\label{sec:inverse_design}
The problem of inverse design aims at finding a set of high-dimensional design parameters (e.g., boundary and initial conditions) for a physical system to optimize a set of specified objectives and constraints. It occurs across many engineering domains such as mechanical, materials, and aerospace engineering \cite{zhang2307artificial}. In the context of CFD, this problem can be formulated as:

\begin{equation} \label{eq:inverse_design}
\begin{split}
\mathbf{u}^*, p^*, \gamma^* = \arg\min_{\mathbf{u},p, \gamma}\mathcal{J}(\mathbf{u},p, \gamma)&, \\
s.t. \quad C(\mathbf{u}, p,\gamma)=0,&
\end{split}
\end{equation}
where $\gamma$ represents the design parameters, $(\mathbf{u}, p)$ contains velocity and pressure in E.q. \eqref{eq:NS}, $\mathcal{J}$ is the design objective, and $C(\mathbf{u}, p,\gamma)=0$ comprises the PDE constraints in E.q. \eqref{eq:NS} with $\gamma$ involved additionally as initial and boundary conditions. This inverse design is closely related to inverse problems in CFD, which takes a similar formulation as in the problem \eqref{eq:inverse_design}. The main difference is that the inverse problems adopt $\mathcal{J}$ as a measure of PDE state estimation error, while the inverse design problems use $\mathcal{J}$ to evaluate a particular design property \cite{zhang2307artificial}. Nevertheless, models and algorithms for both kinds of problems are similar.

The inverse design problem is challenging due to the following factors: (1) it requires accurate and efficient simulation of the PDE systems; (2) the design space is typically high-dimensional with complex constraints; (3) generalization ability to more complex scenarios than observed samples in data-driven inverse design tasks. Along with the rapid progress of ML for PDE simulations, ML for inverse design also gained increasing attention in recent years. Existing work could generally be classified into three categories shown in Fig \ref{fig:inverse_design}. 

\subsubsection{PDE-constrained Methods}
\label{sec:PDE-constrained}
Given the explicit formulation of PDE dynamics, a straightforward approach for inverse design is to optimize the design parameters by minimizing the design objective through the PDE dynamics. This approach is closely related to PINN. Considering all the constraints in PINNs are soft, hPINN imposes hard constraints by using the penalty method and the augmented Lagrangian method \cite{lu2021physics}. Later on, to improve the accuracy and training efficiency of PINNs \cite{yuGradientenhancedPhysicsinformedNeural2022}, gPINN utilizes the gradient information from the PDE residual and incorporates it into the loss function.
Bi-PINN~\cite{hao2022bi} further presents a novel bi-level optimization framework by decoupling the optimization of the objective and constraints to avoid setting the subtle hyperparameter in the unconstrained version of the problem \eqref{eq:inverse_design}. 
\cite{pokkunuru2022improved} proposes a Bayesian approach through a data-driven energy-based model (EBM) as a prior, to improve the overall accuracy and quality of tomographic reconstruction. Towards the inverse problem where the training of PINNs may be sensitive, the data-driven energy-based model (EBM) is introduced as a prior in a Bayesian approach by a recent study on electrical impedance tomography. 

\subsubsection{Data-driven Methods}
\label{sec:Data_driven_inverse}
For scenarios where explicit PDE dynamics are not attainable, inverse design can be performed in a data-driven way.
A representative paradigm is proposed in \cite{allen2022inverse}. It first trains a surrogate model to approximate the PDE dynamics, then optimizes the design parameters through backpropagation of the surrogate model by minimizing the design objective.
The effectiveness of this paradigm in CFD inverse design is verified by manipulating fluid flows and optimizing the shape of an airfoil to minimize drag. 
Regarding the architectures of the surrogate models, GNNs are natural choices as they are effective in characterizing the dependency of the design objective on design parameters over the evolution of PDE dynamics on mesh data structures \cite{zhaoqq2022learning,allen2022inverse}.
In addition, by employing the latent space, 
computational efficiency could be significantly improved
since backpropagation is performed in a much smaller latent dimension and evolution model \cite{wu2022learning}.

Another line of data-driven methods is not limited to the inverse design of PDEs. Instead, these methods are proposed for general inverse design tasks. By including them in this survey, we aim to provide a broader view of the inverse design problems, which may inspire future research on the inverse design of CFD. The main idea of these methods is to directly learn an inverse mapping from the design objective to the design parameters \cite{ardizzone2018analyzing,kruse2021benchmarking}. 
The model variants include 
Invertible Neural Network (INN) \cite{ardizzone2018analyzing}, 
Conditional Invertible Neural Network (cINN) \cite{kruse2021benchmarking}, 
Invertible Residual Network \cite{behrmann2019invertible}, and
Invertible Auto-encoder \cite{teng2019invertible}, and so on. They are compared in \cite{kruse2021benchmarking} and a more recent empirical study is conducted in \cite{ren2020benchmarking}.
A more general setting of inverse design is considered in MINs \cite{kumar2020model}, where the objective function is unknown while only a dataset of pairs of design parameters and objective values are available. It learns an inverse mapping from the objective value to the design parameters such that the design parameters maximize the unknown objective function.

\begin{figure}[t]
\centering
\includegraphics[width=0.48\textwidth]{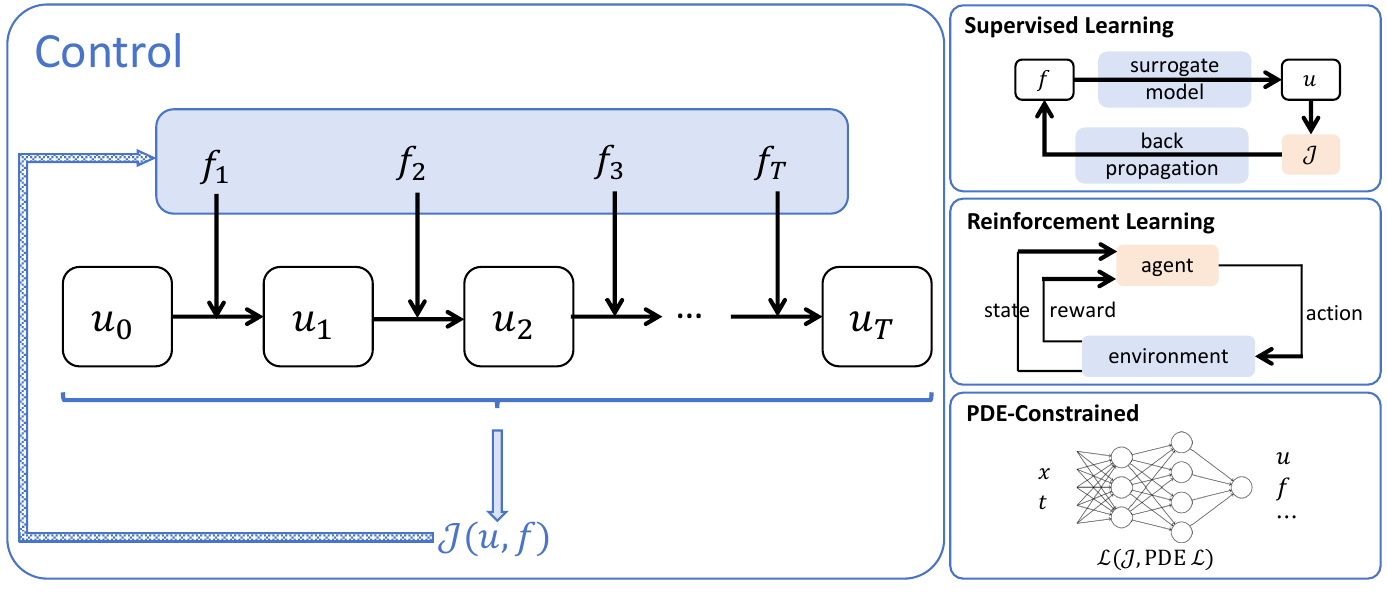}
\vspace{-0.1cm}
\caption{Demonstration of inverse control to control a physical system to achieve a
specific objective with Supervised Learning
Methods, Reinforcement Learning Methods, and PDE-constrained Methods.}
\label{fig:inverse_control}
\vspace{-0.4cm}
\end{figure}

\subsection{Control}
\label{sec:control}
The control problem of PDE systems is also fundamental and has wide applications. The primary goal of control problems is to control a physical system to achieve a specific objective by applying time-varying external forces. The time-varying nature of the external force terms adds complexity to this issue, making it more challenging compared to inverse design. In the context of physical systems constraint to PDEs, the control problem typically involves formulating a control function \( f \) to steer the system:
\begin{align}
        f^* &= \arg\min_{f}\mathcal{J}(f, u), \\
       s.t.\quad \frac{\partial u}{\partial t} &= F\left(u, \frac{\partial u}{\partial x}, \frac{\partial^2 u}{\partial x^2}, \dots\right) +  f(t,x),
\end{align}
where $u$ is the solution, $f$ is the external force term, $F$ is a function and \( \mathcal{J} \) represents the objective of control.

In the last few decades, there have been widely used classical methods for solving PDE control problems including adjoint methods \cite{ mcnamara2004fluid, protas2008adjoint}, Proportional-Integral-Derivative \cite{1580152} and Model Predictive Control \cite{schwenzer2021review}. However, these approaches possess certain drawbacks, including high computational costs and limited applicability. 

Consequently, ML techniques have emerged as a popular method for addressing these issues. In the domain of fluid dynamics \cite{viquerat2022review}, various specific problems such as drag reduction \cite{rabault2019artificial, elhawary2020deep}, conjugate heat transfer \cite{beintema2020controlling, hachem2021deep}, and swimming \cite{ verma2018efficient} have been addressed using ML techniques. We categorize the ML-based methods into three types for the following discussion, which is also shown in Fig \ref{fig:inverse_control}.

\subsubsection{Supervised Learning Methods}
\label{sec:contril_sl}
A significant category of deep learning-based control methods involves training a surrogate model for forward prediction, and then obtaining the optimal control utilizing gradients derived via backpropagation. In \cite{holl2020learning}, researchers introduce a hierarchical predictor-corrector approach for controlling complex nonlinear physical systems across extended periods. In a subsequent study, researchers develop a more direct two-stage method: initially learn the solution operator and subsequently employ gradient descent to determine the optimal control \cite{hwang2022solving}.

\subsubsection{Reinforcement Learning Methods}
\label{sec:contril_rl}
Deep reinforcement learning algorithms are also widely applied to control fluid systems \cite{viquerat2022review, garnier2021review, GARNIER2021104973}. Existing works have extensively utilized a wide range of reinforcement learning algorithms, including Deep Q Learning, Deep Deterministic Policy Gradient, Trust Region Policy Optimization, Soft Actor Critic and Proximal policy optimization \cite{shakya2023reinforcement}. There is also an open-source \texttt{Python} platform DRLinFluids \cite{0103113} that implements numerous reinforcement learning algorithms for controlling fluid systems. The embedded algorithms within the platform include famous model-free RL methods and model-based RL methods. These methods above usually teach an agent to make decisions sequentially to maximize the reward and do not consider physics information directly.

\subsubsection{PDE-constrained Methods}
\label{sec:contril_PDE-Constrained}
In addition to the two types of methods mentioned above, there is another category of algorithms that can find control signals that meet the requirements solely through the form of PDEs without data. These algorithms are all based on PINN \cite{raissi2019physics}. \cite{MOWLAVI2023111731} propose a concise two-stage method. Initially, they train PINN's parameters by addressing a forward problem. Subsequently, they employ a straightforward yet potent line search strategy by evaluating the control objective using a separate PINN forward computation that takes the PINN optimal control as input. Control Physics-Informed Neural Networks (Control PINNs) \cite{barrystraume2022physicsinformed}, on the contrary, is a one-stage approach that learns the system states, the
adjoint system states, and the optimal control signals simultaneously. The first method may be computationally intensive and produce non-physical results due to the indirect relationships in the model. As for the second approach, it offers direct computation of variables and more efficient handling of complex systems, but it may result in large systems of equations.

\subsection{Discussion}

The rapid development of generative models, especially the recent diffusion models \cite{ho2020denoising}, opens a new horizon for the inverse design of CFD. The significant advantage of diffusion models is their ability to effectively sample from high-dimensional distributions, which inspires the generative inverse design paradigm. 
The recent approach SMDP optimizes the inital state of a physical system by moving the system's current state backward in time step by step via an approximate inverse physics simulator and a learned correction function \cite{holzschuh2023solving}.
In contrast, CinDM \cite{wu2024compositional}, which optimizes the energy function captured by diffusion models of observed trajectories, denoises the whole trajectories from a random noise and does not involve a simulator.  A notable feature of  CinDM is that it enables flexible compositional inverse design during inference. Due to the promising performance of diffusion models in various design tasks \cite{watson2023novo,bastek2023inverse}, we believe that they will become one of the mainstream methods for inverse design in the future.

As for the control problems, several new methodologies have emerged recently, employing novel architectural frameworks. \cite{chen2021decisiontransformer} introduces the Decision Transformer, an architecture that frames decision making as a problem of conditional sequence modeling, using a causally masked Transformer to directly output optimal actions. The model conditions on desired returns, past states, and actions, enabling it to generate future actions that achieve specified goals. \cite{wei2024generative} employs a diffusion model to simultaneously learn the entire trajectories of both states and control signals. During inference, they introduce guidance related to the objective \(\mathcal{J}\) along with the prior reweighting technique to assist in identifying control signals closer to optimal.

\section{Applications}
\label{sec:app}

\subsection{Aerodynamics}
\label{sub:Aerodynamics}
In the realm of aerodynamics, CFD is used to simulate and analyze the flow of air over aircraft surfaces, optimizing design for improved performance and efficiency. ML methods have emerged as a transformative force, driving forward the capabilities for more precise simulations and innovative design methodologies.
Mao \textit{et al.}~\cite{mao2020physics} utilize PINNs to approximate the Euler equations, which are crucial for modeling high-speed aerodynamic phenomena. Similarly, Huang \textit{et al.}~\cite{huang2022direct} explores the integration of PINNs with the direct-forcing immersed boundary method, pioneering a novel approach within computational fluid dynamics that enhances the simulation capabilities of boundary interactions in fluid flows. Sharma \textit{et al.}~\cite{sharma2023physics} has developed a physics-informed ML method that integrates neural networks with physical laws to predict melt pool dynamics, such as temperature, velocity, and pressure, without relying on any training data for velocity. This optimization of the PINN architecture significantly enhances the efficiency of model training. Auddy \textit{et al.} \cite{auddy2023grinn} introduces the Gravity-Informed Neural Network (GRINN), a PINN-based framework designed to simulate 3D self-gravitating hydrodynamic systems, showing great potential for modeling astrophysical flows. On the other hand, Shan \textit{et al.}~\cite{shan2023turbulence} apples these networks to turbulent flows involving both attached and separated conditions, significantly improving prediction accuracy for new flow conditions and varying airfoil shapes.
Furthermore, DAIML~\cite{o2022neural} illustrates how data-driven optimization can lead to the creation of highly efficient airfoil shapes for aerial robots, thereby breaking new ground in the optimization of aerodynamic performance. 
And Deng \textit{et al.} \cite{deng2023prediction} employs a transformer-based encoder-decoder network for the prediction of transonic flow over supercritical airfoils, ingeniously encoding the geometric input with diverse information points.
More recently, DIP~\cite{mufti2024shock} showcases ML models' capability to predict aerodynamic flows around complex shapes, enhancing simulation accuracy while reducing computational demands.
These studies collectively underscore the pivotal role of ML in pushing the boundaries of traditional aerodynamics, offering novel solutions that range from refining flow dynamics to achieving superior design efficacy.

\subsection{Combustion \& Reacting Flow}
\label{sub:Combustion}
Applications in turbulent combustion is extensive, encompassing areas such as chemical reactions, combustion modeling, engine performance and combustion instabilities predictions. CFD is utilized to model and study the complex interactions of chemical reactions and fluid dynamics, aiding in the design of efficient and cleaner combustion systems.
Ji \textit{et al.} \cite{Ji_2021} initially explores the efficacy of PINN in addressing stiff chemical kinetic problems governed by stiff ordinary differential equations. The findings demonstrates the effectiveness of mitigating stiffness through the use of Quasi-Steady-State Assumptions.
Besides, Zhang \textit{et al.} \cite{zhang2022multiscale} suggests employing Box-Cox transformation and multi-scale sampling for preprocessing combustion data. It indicates that while the DNN trained on manifold data successfully captures chemical kinetics in limited configurations, it lacks robustness against perturbations.

\subsection{Atmosphere \& Ocean Science}
\label{sub:Atmosphere}
CFD is employed in atmosphere and ocean science to simulate and predict weather patterns, ocean currents, and climate dynamics, enhancing our understanding of environmental systems. Starting with atmospheric applications, Fourcastnet~\cite{pathak2022fourcastnet} enhances forecasts of dynamic atmospheric conditions, such as wind speed and precipitation. Then, PanGu-weather~\cite{bi2023accurate} demonstrates how deep networks with Earth-specific priors effectively capture complex weather patterns. Furthermore, GraphCast~\cite{lam2023learning} predicts weather variables globally at a high resolution in under a minute, showcasing ML's speed and accuracy. Transitioning to oceanic science, Rajagopal \textit{et al.} \cite{rajagopal2023evaluation} excels in forecasting ocean current time series data, which significantly enhances marine navigation and climate research. Moreover, in addressing the pivotal challenge of $CO_2$ trajectory prediction, U-FNO~\cite{wen2022u} leverages advanced ML architectures to tackle complex multiphase problems, representing a significant step forward in environmental modeling. Complementing this, Fourier-MIONet~\cite{jiang2023fourier} offers innovative solutions to computational challenges in 4D numerical simulations. In essence, it marks a significant leap in atmosphere modeling, enhancing the accuracy and resolution of environmental models. This evolution not only enriches our understanding of complex natural phenomena but also opens new avenues for addressing global environmental challenges.

\subsection{Biology Fluid}
\label{sub:Biology}
CFD is also applied in biological fluid dynamics to model and analyze the behavior of fluids within biological systems, such as blood flow in arteries, contributing to advancements in medical research and healthcare. Yin \textit{et al.}~\cite{yin2022simulating} utilizes DeepONet to simulate the delamination process in aortic dissection. This approach allows for the accurate prediction of differential strut distributions, marking a significant advancement in modeling the biomechanics of aortic dissection.
Besides, Vooter \textit{et al.}~\cite{voorter2023improving} employs PINNs to enhance the analysis of multi-b-value diffusion and extracts detailed biofluid information from MRI data, focusing on the microstructural integrity of interstitial fluid and microvascular images. This innovation not only improves the quality of biomedical imaging but also contributes to the broader understanding of microvascular flows and tissue integrity.
Then, Shen \textit{et al.}~\cite{shen2023multiple} explores the capabilities of multi-case PINNs in simulating biomedical tube flows, particularly in scenarios involving varied geometries. By parameterizing and pre-training the network on different geometric cases, the model can predict flows in unseen geometries in real-time. Collectively, the advancements in surrogate modeling and physics-informed analysis represent critical steps forward in bioengineering, medical diagnostics, and treatment planning, showcasing the potential of ML to revolutionize intervention strategies in biofluid dynamics.

\subsection{Plasma}
\label{sub:Plasma}
CFD is used in plasma physics to simulate and analyze the behavior of ionized gases, aiding in the development of applications such as nuclear fusion, space propulsion, and advanced materials processing. Simulations involving high-dimensional and high-frequency plasma dynamics, are particularly well-suited for the application of ML methods. Zhong \textit{et al.} \cite{Zhong_2022} introduces two innovative networks to address specific challenges. The first, CS-PINN, is designed to approximate solution-dependent coefficients within plasma equations. The second, RK-PINN, integrates the implicit Runge–Kutta formalism to facilitate large-time step predictions for transient plasma behavior.
Gopakumar \textit{et al.} \cite{gopakumar2023fourier} have illustrated that FNO can predict the magneto-hydrodynamic models governing plasma dynamics with a speed that is six orders of magnitude faster than traditional numerical solvers, while still achieving a notable accuracy.
Furthermore, Kim \textit{et al.} \cite{kim2024highest} presents an innovative 3D field optimization approach that leverages ML and real-time adaptability to overcome the instability of the transient energy burst at the boundary of plasmas.

\subsection{System Identification and Symbolic Regression}
\label{sub:symbolic}
Identifying and recovering the governing equations for dynamical systems are important modeling problems where one approximates the underlying equation of motion using data-driven methods. Bongard and Lipson \cite{bongard2007automated} introduce the symbolic regression for modeling the variables from time series. The SINDy algorithm \cite{Brunton_2016} identifies the governing equations by learning a sparse representation of the dynamical system from a dictionary of candidate functions, usually consisting of trigonometric or polynomial functions. Sparse optimization methods for identifying the model from spatio-temporal data are developed in \cite{schaeffer2017learning}, where differential operators are included in the dictionary. The governing equation is obtained using the LASSO method, which offers explicit error bounds and recovery rates. The sparse optimization techniques are later integrated with PINNs \cite{chen2021physics} to obtain models from scarce data. 

More recently, symbolic regression techniques are combined with modern architectures to balance performance and explanability. In \cite{becker2023predicting}, a sequence-to-sequence transformer model is used to directly output the underlying ODEs in prefix notation. The FEX method \cite{jiang2023finite} uses deep reinforcement learning to approximate the governing PDEs with symbolic trees. Although these methods haven't been directly applied to CFD, the governing equations obtained using these methods can be used for future simulation when the underlying model is unknown.

\subsection{Reduced Order Modeling}
\label{sub:ROMs}
Reduced-order modeling (ROM) involves identifying a space that captures the dynamics of a full model with significantly fewer degrees of freedom (DOFs), such as Dynamic Mode Decomposition (DMD) and the Koopman method. Traditionally, these methods heavily rely on domain knowledge and are typically applied to simple examples. The development of ML enables the learning of more complex and realistic systems. For instance, combining ML architectures like CNNs~\cite{leask2021modal} and transformers~\cite{geneva2022transformers} with the Koopman method has shown positive results.
Incorporating symmetries (or physics system invariant) into the model architecture~\cite{kneer2021symmetry} is a solid technical improvement, reducing the amount of training data required and allowing the learned ROM to obey the governing rules explicitly.

Besides serving as a surrogate model, a learned ROM can also be combined with numerical solvers via Galerkin projection to form a hybrid approach. Specifically, the governing equations can be projected onto a reduced space to be solved with less computational cost. For example, Arnold \textit{et al.} \cite{arnold2022large} apply adaptive projection-based reduced modeling for combustion-flow problems in gas turbine engines. By using the first 1\% of the full model simulation, they identify a reduced space that allowed the simulation to achieve a 100x speedup using ROM. Wentland \textit{et al.}~\cite{wentland2023scalable} subsequently extend this approach to general multi-phase flow simulations.

\section{Challenges \& Future works}
\label{sec:chall}

\subsection{Multi-Scale Dynamics}
\label{subsec:ms}
The challenge of multi-scale modeling lies in accurately capturing the interactions across vastly different scales, from microscopic molecular motions to macroscopic flow behaviors, within the constraints of limited high-fidelity data and computational resources. Fortunately, ML has been pivotal in bridging the gap caused by limited high-fidelity data availability. The challenge is compounded by the intrinsic complexity of multi-scale systems, where phenomena at different scales can influence each other in non-linear and often unpredictable ways. For instance, microscopic molecular dynamics can have significant impacts on macroscopic properties such as viscosity and turbulence in fluid flows.

\noindent\textbf{Representative works.} Vlachas \textit{et al.}~\cite{vlachas2022multiscale} have leveraged auto-encoders to establish a connection between fine- and coarse-grained representations, subsequently evolving the dynamics within the latent space through RNNs.
Furthermore, Lyu \textit{et al.}~\cite{lyu2023multi} develop a multi-fidelity learning approach using the FNO that synergizes abundant low-fidelity data with scarce high-fidelity data within a transfer learning framework. This multi-level data integration is similarly echoed in Cascade-Net~\cite{mi2023cascade}, which hierarchically predicts velocity fluctuations across different scales, ensuring the energy cascade in turbulence is accurately represented.
Innovations also extend to GNNs, where the BSMS-GNN model~\cite{cao2023efficient} introduces a novel pooling strategy to incorporate multi-scale structures. 
Concurrently, the MS-MGN~\cite{fortunato2022multiscale} adapts GNNs for multi-scale challenges by manually drawing meshes. 
More recently, Fang \textit{et al.}~\cite{fang2023solving} utilize the Gaussian process conditional mean for efficient predictions at massive collocation points. Additionally, SineNet~\cite{zhang2024sinenet} employs a sequence of U-shaped network blocks to refine high-resolution features progressively across multiple stages. 

Another roadmap focuses on altering the solution space's degree of freedom and the topology during mesh refinement to enhance simulation accuracy and computational efficiency. For example, LAMP~\cite{wu2023learning} utilizes a reinforcement learning algorithm to dictate the h-adaptation policy, facilitating dynamic changes in the topology of meshes.
Furthermore, CORAL~\cite{serrano2023operator} revolutionizes mesh adaptation by removing constraints on the input mesh. 

\noindent\textbf{Promising future.} One promising direction is the development of hybrid models that seamlessly combine data-driven approaches with traditional physics-based simulations, enhancing their ability to generalize across different scales and scenarios. The continuous improvement of transfer learning techniques will also play a crucial role, enabling models to leverage knowledge from related problems and datasets to improve performance with limited high-fidelity data.
Moreover, the exploration of novel architectures will further advance the capability to capture complex interactions across scales. These architectures can be enhanced with more sophisticated pooling and aggregation strategies, as well as improved interpretability to ensure that the learned models adhere to known physical laws.
Additionally, advancements in computational hardware, such as the use of specialized processors and distributed computing frameworks, will enable the execution of more complex and large-scale simulations.

\subsection{Explicit Physical Knowledge Encoding}
\label{subsec:phys_encode}

Another primary challenges is effectively incorporating the fundamental physical laws governing fluid dynamics from diverse sources explicitly into a coherent high-dimensional and nonlinear framework. Explicitly integrating physical knowledge differs from PINNs in that the former directly incorporates physical laws and constraints into the model, while PINNs embed these laws within the neural network's loss function to guide the learning process. 

\noindent\textbf{Representative works.} Raissi \textit{et al.}~\cite{raissi2020hidden} leverage the NS equations to inform the learning process, ensuring that the dynamics of fluid flow are accurately captured in scientifically relevant scenarios. Further advancements are seen in approaches like FINN~\cite{karlbauer2022composing}, which merges the learning capabilities of ANNs with the physical and structural insights derived from numerical simulations.
More recently, Sun \textit{et al.}~\cite{rao2023encoding} have introduced a framework that integrates a specific physics structure into a recurrent CNN, aiming to enhance the learning of spatio-temporal dynamics in scenarios with sparse data.

As for encoding boundary and initial conditions, Sun \textit{et al.}~\cite{sun2020surrogate} introduce a physics-constrained DNN and incorporate boundary conditions directly into the learning process. 
BOON~\cite{saad2022guiding} modifies the operator kernel to ensure boundary condition satisfaction.
Besides, Rao \textit{et al.}~\cite{rao2023encoding} take a different approach by encoding the physical structure directly into a recurrent CNN. BENO~\cite{wang2024beno} presents a boundary-embedded neural operator that integrates complex boundary shapes and inhomogeneous boundary values into the solution of elliptic PDEs.
Neural IVP~\cite{finzi2023stable} offers a solver for initial value problems based on ODEs. By preventing the network from becoming ill-conditioned, Neural IVP enables the evolution of complex PDE dynamics within neural networks.

\noindent\textbf{Promising future.} Future meaningful research directions include the development of more novel implicit network architectures. These architectures should be designed to embed physical knowledge seamlessly. Furthermore, combining manifold learning and graph relationship learning techniques in ML can help extract underlying physical relationships and laws. This approach aims to enhance the ability of ML models to understand and incorporate complex physical systems, leading to more accurate predictions.

\subsection{Multi-physics Learning \& Scientific Foundation Model}
\label{subsec:mpp}
A primary objective for scientific ML is to develop methods that can generalize and extrapolate beyond the training data. 
Surrogate models typically perform well only under the working conditions or geometries for which they are trained. Specifically, PINNs often solve only a single instance of PDE, while Neural Operator generalizes only to a specific family of parametric PDEs. Similarly, ML-assisted approaches, such as those for closure modeling, tend to be limited by their working conditions or the shapes of walls during training.

\noindent\textbf{Representative works.} Lozano \textit{et al.}~\cite{lozano2023machine} involve training multiple ML wall models as candidates for different flow regimes, followed by training a classifier to select the most suitable candidate for the current conditions.
In-Context Operator Network (ICON)~\cite{yang2023context,yang2023prompting,yang2024pde}, leverages pairs of physics fields before and after numerical time steps as contexts. This model learns to solve the time steps based on these contexts, showcasing an ability to handle \textit{different time scales} and generalize across \textit{different types of PDE operators}. 
PROSE~\cite{liu2023prose,sun2024towards} combines data and equation information through a multimodal fusion approach for simultaneous prediction and system identification, and demonstrates \textit{zero-shot} extrapolation to \textit{different data distribution}, \textit{unseen physical features}, and \textit{unseen equations}. 
FMint~\cite{song2024fmint} combines the precision of human-designed algorithms with the adaptability of data-driven methods, and is specifically engineered to achieve \textit{high-accuracy simulation} of dynamical systems. 
Unisolver~\cite{hang2024unisolver} integrates all PDE components available, including equation symbols, boundary information and PDE coefficients. By separately process domain-wise (e.g. equations) and point-wise information (e.g. boundaries), Unisolver shows generalization to out-of-distribution parameters.  

Another line of works focus on transfer learning techniques, where a pretrained model is finetuned to align with downstream tasks. 
Subramanian \textit{et al.}~\cite{subramanian2024towards} use a transfer learning approach, showing models pre-trained on multi-physics can be adapted to various downstream tasks.
The multi-physics pre-train (MPP)~\cite{mccabe2023multiple} approach, akin to many brute force LLMs, seeks to establish a foundational model applicable across diverse physics systems.
OmniArch~\cite{chen2024building} pre-trains on all 1D/2D/3D PDE data using auto-regressive tasks and fine-tunes with physics-informed reinforcement learning to align with physical laws, which excels in few-shot and zero-shot learning for new physics.
DPOT~\cite{hao2024dpot} introduces a novel auto-regressive denoising pre-training approach that enhances the stability and efficiency of pre-training.

\noindent\textbf{Promising future.} One promising future direction is to design the network to simultaneously handle different complex geometries. This requires having a network capable of processing heterogeneous data, as well as a large collection of high-quality (real \& synthetic) training data. Besides, while pretrained LLMs are not directly suitable for scientific computing tasks, incorporating their huge pretrained knowledge base would be beneficial, especially in the data-scarce regime. Additionally, small language models equipped with scalable training strategies~\cite{hu2024minicpm} can provide an effective and efficient approach.

\subsection{Automatic Data Generation \& Scientific Discovery}
The success of all the aforementioned applications in this review heavily depends on the size and coverage of the training dataset. This is especially true for multi-physics models as in Sec.~\ref{subsec:mpp}, as demonstrated by emerging effects in LLMs~\cite{wei2022emergent}. Unlike textual or visual data readily available online, just like many scientific domains, CFD data is characterized by a large number of samples due to complex system parameter combinations, spanning a wide variety of different models, and typically incurring significantly high costs to obtain. This combination presents a significant challenge in generating a sufficiently large and diverse dataset for the above purposes.
In the previous Sec.~\ref{subsec:ms} and Sec.~\ref{subsec:phys_encode}, we mention incorporating symmetries and physics knowledge to decrease the dependence on training dataset size. However, automatically and efficiently guiding the ML model to generate data still poses challenges.

\noindent\textbf{Representative works.} Generative modeling, notably diffusion models~\cite{ho2020denoising}, has recently emerged as a promising direction for high-quality generation in computer vision.
Conditioned diffusion models~\cite{batzolis2021conditional} provide more control during the generation of designed dataset samples and is a promising candidate for automatic data generation.
Diffusion models have been naturally extended to CT, medical imaging, and MRI~\cite{liu2023artificial}, turbulence~\cite{gao2024bayesian},and molecular dynamics~\cite{arts2023two}, and more.

\noindent\textbf{Promising future.}
Automated experimentation (auto-lab) has become a promising pipeline for automatic data generation and scientific discovery. By leveraging trained surrogate models (the verifier), auto-lab trains an additional ML model to propose trials (the proposer), which can then be efficiently filtered by the verifier to retain only those with a high success rate. Real experiments or high-fidelity simulations are carried out only on filtered trials. The obtained results progressively enrich the dataset and retrain the ML models to highlight trial directions with higher success rates. This pipeline tends to automate and scale up traditional experiments and has been applied in materials science~\cite{merchant2023scaling,du2023m}, metamaterials~\cite{ha2023rapid}, protein structure~\cite{jumper2021highly}, robotics~\cite{pyzer2022accelerating} and more. Similar works can be promising in the field of CFD and related areas.

\section{Conclusion}
In conclusion, this paper has systematically explored the significant advancements in leveraging ML for CFD. We have proposed a novel classification approach for forward modeling and inverse problems, and provided a detailed introduction to the latest methodologies developed in the past five years. We also highlight the promising applications of ML in critical scientific and engineering domains. Additionally, we discussed the challenges and future research directions in this rapidly evolving domain. Overall, it is evident that ML has the potential to significantly transform CFD research.

\bibliographystyle{IEEEtran}
\bibliography{sn-bibliography}

\clearpage
\section{Appendix}
Here are refined explanations for each PDE, adding more detail to their descriptions. In addition, the discussion of flow problems introduces scenarios such as Ahmed-Body flow, crucial for automotive aerodynamics research, emphasizing drag reduction and flow separation. 

\subsection{Advection Equation}
Advection equation models the transport of a scalar quantity with the fluid flow at a velocity $u$: 
\begin{equation*}
    \frac{\partial \phi}{\partial t} + u \frac{\partial \phi}{\partial x} = 0
\end{equation*}

\subsection{Allen-Cahn Equation}
Allen-Cahn equation describes the phase field dynamics of multi-phase systems, capturing the evolution of interfaces between different phases with the parameter $\epsilon$ controlling the interface width:
\begin{equation*}
    \frac{\partial \phi}{\partial t} = \epsilon \Delta \phi - \frac{1}{\epsilon} \phi (\phi^2 - 1)
\end{equation*}

\subsection{Anti-Derivative Equation}
Anti-Derivative equation models the mathematical property where the derivative of an integral of a function returns the original function:
\begin{equation*}
    \frac{d}{dx} \left(\int \phi \, dx\right) = \phi
\end{equation*}

\subsection{Bateman–BurgersEquation}
Bateman–Burgers equation combines the effects of viscous diffusion and nonlinear convection in fluid dynamics:
\begin{equation*}
    \frac{\partial u}{\partial t} + u \frac{\partial u}{\partial x} = \nu \frac{\partial^2 u}{\partial x^2}
\end{equation*}

\subsection{Burgers Equation}
Burgers is a fundamental equation in fluid mechanics and nonlinear acoustics that simplifies the dynamics of fluid flows with viscosity. It is particularly useful in the study of shock wave formation and turbulence modeling:
\begin{equation*}
    \frac{\partial u}{\partial t} + u \frac{\partial u}{\partial x} = \nu \frac{\partial^2 u}{\partial x^2}
\end{equation*}

\subsection{Diffusion Equation}
Diffusion equation is for describing how substances spread through space over time due to diffusion. It's a basic model for heat conduction, mass transport in porous media, and other diffusion processes:
\begin{equation*}
    \frac{\partial \phi}{\partial t} = D \Delta \phi
\end{equation*}

\subsection{Duffing Equation}
Duffing equation is a non-linear differential equation models the dynamics of oscillators with non-linear restoring forces, which can exhibit both periodic and chaotic behavior:
\begin{equation*}
    \frac{d^2 x}{dt^2} + \delta \frac{dx}{dt} + \alpha x + \beta x^3 = \gamma \cos(\omega t)
\end{equation*}

\subsection{Eikonal Equation}
Eikonal equation is essential in optics and computational geometry for modeling the propagation of wavefronts: 
\begin{equation*}
    \|\nabla \phi\| = F(x)
\end{equation*}

\subsection{Elastodynamic Equation}
Elastodynamic equation models the behavior of elastic materials subjected to external stresses and strains:
\begin{equation*}
    \rho \frac{\partial^2 u}{\partial t^2} - \mu \Delta u - (\lambda + \mu) \nabla(\nabla \cdot u) = 0
\end{equation*}

\subsection{Euler Equation}
Euler equation governs the motion of an inviscid fluid, emphasizing the conservation of mass and momentum:
\begin{equation*}
    \frac{\partial \rho}{\partial t} + \nabla \cdot (\rho \vec{u}) = 0
\end{equation*}

\subsection{Gray-Scott Equation }
Gray-Scott equations model the dynamics of reaction and diffusion processes in chemical kinetics, specifically the interaction between two chemical species undergoing reaction and diffusion:
\begin{equation*}
    \frac{\partial u}{\partial t} = -uv^2 + F(1-u) + D_u \Delta u
\end{equation*}

\subsection{Heat Equation}
Heat equation describes how temperature changes over time within a given region, accounting for thermal diffusion:
\begin{equation*}
    \frac{\partial T}{\partial t} = \kappa \Delta T
\end{equation*}

\subsection{Korteweg-de Vries Equation}
Korteweg-de Vries equation models the propagation of solitary waves in shallow water and is crucial in the study of nonlinear wave phenomena in various physical contexts:
\begin{equation*}
    \frac{\partial u}{\partial t} + u \frac{\partial u}{\partial x} + \delta \frac{\partial^3 u}{\partial x^3} = 0
\end{equation*}

\subsection{Kuramoto-Sivashinsky Equation}
Kuramoto-Sivashinsky equation captures the behavior of instabilities and chaotic patterns in systems such as laminar flame fronts and fluid interfaces:
\begin{equation*}
    \frac{\partial u}{\partial t} + \nu \frac{\partial^4 u}{\partial x^4} + \frac{\partial^2 u}{\partial x^2} + u \frac{\partial u}{\partial x} = 0
\end{equation*}

\subsection{Laplace Equation}
Laplace equation is fundamental in electrostatics, fluid mechanics, and geometric theory, describing potential fields in the absence of free charges or sources:
\begin{equation*}
    \Delta \phi = 0
\end{equation*}

\subsection{Poisson Equation}
Poisson equation is a generalization of the Laplace equation, incorporating a source term, $f$, which describes the potential field $\phi$ in response to a given source distribution:
\begin{equation*}
    \Delta \phi = f
\end{equation*}

\subsection{Reynold Averaged Navier-Stokes Equation}
Reynold Averaged Navier-Stokes equation models turbulent flows by averaging the NS equations over time. It is a crucial tool in fluid dynamics for designing and analyzing systems where turbulence plays a significant role, simplifying the complex interactions of turbulent flow:
\begin{equation*}
    \rho \left(\frac{\partial \vec{u}}{\partial t} + \vec{u} \cdot \nabla \vec{u}\right) = -\nabla p + \mu \nabla^2 \vec{u} + \rho \vec{f}
\end{equation*}

\subsection{Reaction-Diffusion Equation}
Reaction-Diffusion equation combines chemical reaction dynamics with diffusion processes to describe patterns such as stripes, spirals, and chaos that emerge in chemical and biological systems:
\begin{equation*}
    \frac{\partial \phi}{\partial t} = D \Delta \phi + R(\phi)
\end{equation*}

\subsection{Schr\''{o}dinger Equation}
Schr\''{o}dinger equation describes how the quantum state of a physical system changes over time:
\begin{equation*}
    i\hbar \frac{\partial \psi}{\partial t} = -\frac{\hbar^2}{2m} \Delta \psi + V \psi
\end{equation*}

\subsection{Shallow Water Equation}
Shallow Water equations describe fluid flow in shallow waters, where the horizontal dimensions are much greater than the vertical dimension:
\begin{equation*}
    \frac{\partial h}{\partial t} + \nabla \cdot (h \vec{u}) = 0
\end{equation*}

\subsection{Wave Equation}
Wave equation models the propagation of various types of waves, such as sound, light, and water waves through different mediums:
\begin{equation*}
    \frac{\partial^2 \phi}{\partial t^2} = c^2 \Delta \phi
\end{equation*}

\subsection{Airfoil Flow}
Airfoil flow, governed by the Navier-Stokes equations, serves to analyze lift and drag in aeronautical engineering, spanning subsonic to supersonic conditions. 
In more complex geometries, Beltrami flow, characterized by:
\begin{equation*}
    \nabla \times u = \lambda u
\end{equation*}
which explores fluid dynamics where vorticity aligns with the velocity, offering insights into vortex-dominated flows.

\subsection{Cavity Flow}
Cavity flow, CylinderFlowinder flow, and Dam flow are all driven by the NS equations under specific no-slip boundary conditions.
Darcy flow describes the flow of fluid through porous media as: 
\begin{equation*}
    \nabla \cdot (k \nabla p) = S
\end{equation*}

\subsection{Kovasznay Flow}
Kovasznay flow is a specific solution of the NS equations for low Reynolds number flows.

\subsection{Kolmogorov Flow}
Kolmogorov flow is governed by the NS equations with an external forcing term, typically sinusoidal.
Rayleigh-B\'{e}nard flow investigates convection patterns in coupled heat transfer and fluid flow equations.

\subsection{Transonic Flow}
Transonic flow is critical in aerospace engineering, focusing on the behavior of airflows that include regions of subsonic flow around objects.


\end{document}